%% file: tmlr.tex
\documentclass[10pt]{article} 
\usepackage[preprint]{tmlr}

\input{math_commands.tex}

\usepackage{hyperref}
\usepackage{url}
\usepackage{graphicx}
\usepackage{listingsutf8} 
\usepackage{amsthm}

\usepackage{booktabs} 
\usepackage{listings}

\usepackage{tcolorbox}
\usepackage{array}
\usepackage{longtable}

\usepackage{ragged2e}
\newtcolorbox{promptbox}{
    colback=gray!10,
    colframe=black, 
    coltitle=black, 
    coltext=black,
    fonttitle=\bfseries
}

\usepackage{listings}

\lstdefinelanguage{json}{
    basicstyle=\ttfamily\footnotesize,
    keywordstyle=\bfseries\color{blue},
    stringstyle=\color{red},
    morestring=[b]",
    morecomment=[l]{//},
    commentstyle=\color{gray}
}

\title{DeepProtein: Deep Learning Library and Benchmark\\
for Protein Sequence Learning}

\author{\name Jiaqing Xie \email jiaxie@student.ethz.ch \\
      \addr Department of Computer Science\\
      ETH Zurich
      \AND
      \name Tianfan Fu \email futianfan@gmail.com\\
      \addr National Key Laboratory for Novel Software Technology, School of Computer Science \\
      Nanjing University}

\newcommand{\mname}{\texttt{DeepProtein}}
\def\month{MM}  
\def\year{YYYY} 
\def\openreview{\url{https://openreview.net/forum?id=XXXX}} 

\begin{document}

\maketitle

\begin{abstract}
Deep learning has deeply influenced protein science, enabling breakthroughs in predicting protein properties, {higher-order structures}, and molecular interactions. This paper introduces \mname, a comprehensive and user-friendly deep learning library {tailored} for protein-related tasks. {It enables researchers to seamlessly address protein data with cutting-edge deep learning models. To assess model performance, we establish a benchmark evaluating different deep learning architectures across multiple protein-related tasks,} including protein function prediction, subcellular localization prediction, protein-protein interaction prediction, and protein structure prediction. {Furthermore, we introduce DeepProt-T5, a series of fine-tuned Prot-T5-based models that achieve state-of-the-art performance on four benchmark tasks, while demonstrating competitive results on six of others.} Comprehensive documentation and tutorials are available which could ensure accessibility and support reproducibility. Built upon the widely used drug discovery library DeepPurpose, \mname~is publicly available at \url{https://github.com/jiaqingxie/DeepProtein}.
\end{abstract}

\section{Introduction}

Understanding the representation of proteomics is vital in developing traditional biological and medical progress~\citep{wu2022cosbin,fu2024ddn3}, multi-omics genomics~\citep{lu2022cot,chen2021data},  and curing human diseases~\citep{lu2024uncertainty,chen2024uncertaintyquantificationclinicaltrial}. Being the working house of the cell, it provides many functions that support human daily life, such as catalyzing biochemical reactions that occur in the body as a role of enzymes and providing helpful immune responses against harmful substances that act as immunoglobulin~\citep{wu20241596}. 
Under the necessity of analyzing those useful proteins, several related protein databases are available to researchers~\citep{berman2000protein, bairoch2000swiss, uniprot2015uniprot, ponten2008human}. Apart from the 2D database, some recent 3D Protein Database used AlphaFold 2.0~\citep{jumper2021highly} is important to better assist in learning those representations in 3d-dimensional space. The success of AlphaFold 2.0 has sparked a significant increase in interest in using machine learning techniques for protein learning tasks, of which the goal is to improve our understanding of proteins' biochemical mechanisms.

Deep learning has revolutionized protein science, driving significant advancements in various protein-related tasks. These include protein-protein interactions \citep{gainza2020deciphering}, protein folding \citep{jumper2021highly, lu2022protein, panou2020deepfoldit, chen2016profold}, protein-ligand interactions \citep{li2021structure, xia2023leveraging}, and protein function and property prediction \citep{gligorijevic2021structure, sevgen2023prot}. The development of deep neural architectures has played a crucial role in these tasks, with approaches leveraging both \textit{sequence-based} and \textit{structure-based} models. \textit{Sequence-based} models, such as convolutional neural networks (CNNs) \citep{shanehsazzadeh2020transfer} and transformers, have shown strong performance in protein learning tasks. The TAPE Transformer \citep{rao2019evaluating} and pre-trained transformer models such as ProtBERT \citep{brandes2022proteinbert} have demonstrated the effectiveness of self-supervised learning in capturing protein sequence representations. Beyond sequence-based methods, \textit{structure-based} deep learning has gained traction with graph neural networks (GNNs), which leverage 3D structural information to enhance structural property \citep{jing2020learning, zhang2022protein}. {Recently, graph transformers have emerged as a powerful alternative, combining the advantages of transformers (global attention) and message-passing neural networks (sparse attention) to model protein structures more effectively \citep{yuan2022alphafold2, gu2023hierarchical}.}

While transformers have been considered state-of-the-art in previous benchmarks \citep{xu2022peer}, comprehensive comparisons between CNN, transformer, GNN, and other advanced architectures remain under-explored. This gap motivates us to systematically integrate and evaluate these methods in our benchmark. {Furthermore, pretraining strategies have been prevailing in protein science, which have utilized the large-scale unlabeled protein data to improve downstream performance \citep{lu2024drugclip, yue2024biomamba}. With the advent of large foundation models, protein properties can now be inferred through prompt engineering, such as BioMistral \citep{labrak2024biomistral}, BioT5/BioT5+ \citep{pei2023biot5, pei2024biot5+}, and ChemLLM \citep{zhang2024chemllm}. Both advancements in molecule pretraining and question-answering language models in molecules brought more possibilities in the field of protein engineering.}

\begin{table}[!htb]
\centering
\caption{Comparison of benchmark studies on protein sequence learning. TDC provides AI-ready datasets but does not contain protein learning benchmarks (denoted $\mathbf{\diamond}$). 
}
\label{table:method}
\resizebox{\columnwidth}{!}{
\begin{tabular}{lcccccc}
\toprule[1.0pt]
Datasets  & DeepPurpose & FLIP  & TAPE  & PEER  & TDC (data only) & \mname \\ 
\hline  
References & \citep{huang2020deeppurpose} & \citep{dallago2021flip} & \citep{rao2019evaluating} & \citep{xu2022peer} & \citep{huang2021therapeutics} & ours \\ \midrule  
Fluorescence & $\times$ &  \checkmark & \checkmark & \checkmark &  $\times$  & \checkmark  \\ 
$\beta$-lactamase & $\times$ & $\times$ & $\times$ & \checkmark & $\times$ & \checkmark \\
Solubility & $\times$ & $\times$ & $\times$ & \checkmark &  $\times$  & \checkmark  \\
Stability & $\times$ & \checkmark & \checkmark & \checkmark &  $\times$  & \checkmark\\
Subcellular (Binary) & $\times$ & $\times$ & $\times$ & \checkmark & $\times$ & \checkmark \\
PPI Affinity & $\times$ & $\times$ & $\times$ & \checkmark & $\times$ & \checkmark \\
Yeast PPI & $\times$ & $\times$ & $\times$ & \checkmark & $\times$ & \checkmark \\
Human PPI & $\times$ & $\times$ & $\times$ & \checkmark & $\times$ & \checkmark \\
IEDB & $\times$ & $\times$ & $\times$ & $\times$ & $\mathbf{\diamond}$ & \checkmark \\
PDB-Jespersen & $\times$ & $\times$ & $\times$ & $\times$ & $\mathbf{\diamond}$ & \checkmark \\
SAbDab-Liberis & $\times$ & $\times$ & $\times$ & $\times$ & $\mathbf{\diamond}$ & \checkmark \\
TAP & $\times$ & $\times$ & $\times$ & $\times$ & $\mathbf{\diamond}$ & \checkmark \\
SAbDab-Chen & $\times$ & $\times$ & $\times$ & $\times$ & $\mathbf{\diamond}$ & \checkmark \\
CRISPR-Leenay & $\times$ & $\times$ & $\times$ & $\times$ & $\mathbf{\diamond}$ & \checkmark \\ 
{Fold} & $\times$ & $\times$ & $\times$ & \checkmark & $\times$  & \checkmark\\ 
{Secondary Structure} & $\times$ & $\times$ & \checkmark  & \checkmark & $\times$  & \checkmark\\  
\bottomrule[1.0pt]
\end{tabular}}
\end{table}

\noindent\textbf{Challenges.}
Previous benchmarks related to molecular learning have offered valuable insights regarding their respective libraries and implementation interfaces. DeepPurpose\footnote{\url{https://github.com/kexinhuang12345/DeepPurpose}} \citep{huang2020deeppurpose} has provided an interface that implements the task with a majority of drug discovery tasks, which only has protein-protein interaction and protein function prediction implemented. 
Datasets on proteins are lacking as well. TorchProtein\footnote{\url{https://github.com/DeepGraphLearning/PEER_Benchmark} }~\citep{xu2022peer}, also named as PEER, implemented most of the tasks in the protein field. In terms of models, the focus has largely been on \textit{sequence-based} methods: Convolutional Neural Networks (CNNs), Transformers, and ESM architectures. {This suggests that there are still many \textit{structure-based} methods (GNN) or pre-trained protein language models available (such as ProtBert or Prot-T5) for consideration.} Furthermore, PEER's interface is not user-friendly without prior domain knowledge in graphs and biochemistry. This presents an opportunity to improve the existing interface regarding simplicity and comprehensibility.

\noindent\textbf{Solutions.}
To address these challenges, in this paper, we propose \mname, which aims to benchmark mainstream and cutting-edge deep learning models on a wide range of AI-solvable protein sequence learning tasks. 
We investigate the performance of various deep learning models on a wide range of protein sequence learning tasks. 
We analyze each method's advantages and disadvantages when performing each task (working as the explainer for each task). We have provided user-friendly and well-wrapped interfaces to facilitate domain experts' research. 

\noindent\textbf{Contribution.}
Our key contributions are summarized as: 
\begin{itemize}
\item \textbf{Comprehensive Benchmarking}: {We curate a benchmark to evaluate the performance of eight coarse grained deep learning architectures, including CNNs, CNN-RNNs, RNNs, transformers, graph neural networks, graph transformers, pre-trained protein language models, and large language models. This benchmark covers eight protein learning tasks, including protein function prediction, protein localization prediction, protein-protein interaction prediction, antigen epitope prediction, antibody paratope prediction, CRISPR repair outcome prediction, antibody developability prediction, and protein structure prediction. Our benchmark demonstrates the strength, scalability and limitation of the mentioned approach, respectively.}  
\item \textbf{User-friendly Library}: We develop \mname, a specialized deep learning library that integrates these neural network architectures for protein-related tasks. \mname~offers a simple, command-line interface for running models on all supported tasks, making it accessible to researchers with minimal deep learning expertise.
\item \textbf{Enhanced Accessibility}: {We provide comprehensive documentation, tutorials, and pre-configured pipelines. Inherited from DeepPurpose \citep{huang2020deeppurpose}, our library ensures seamless integration with existing protein frameworks or personalized protein databases, and enables reproducibility.} 
\item {\textbf{Fine-Tuned Models – \texttt{DeepProt-T5}}: We have released our fine-tuned Prot-T5-XL models for each task,  which is available on HuggingFace. The model family is called DeepProt-T5. These models achieve either state-of-the-art or competitive performance across our DeepProtein benchmark, so there is no need for the redundant retraining process, making model deployment much more efficient and convenient.}

\end{itemize}

\section{Related Works} 
Benchmarks and libraries are crucial in AI-based therapeutic science, e.g., multi-omics data~\citep{lu2018multi}, protein learning~\citep{xu2022peer}, small-molecule drug discovery~\citep{gao2022samples,zheng2024structure,xu2024smiles}, and drug development (clinical trial)~\citep{chen2024trialbench,wang2024twin}. They provide standardized metrics for evaluating the performance of various algorithms and models. These benchmarks enable researchers to compare different approaches systematically, ensuring reproducibility and reliability of results. 

In this section, we briefly discuss the benchmark studies in this area. 
Proteins are vital in drug discovery because they often serve as the primary targets for therapeutic agents, influencing disease mechanisms and biological pathways. Additionally, proteins play key roles in various cellular processes, making them essential for identifying potential drug candidates and biomarkers in the drug development pipeline. 
A couple of protein learning benchmarks are developed, including PEER~\citep{xu2022peer}, DeepPurpose~\citep{huang2020deeppurpose}, FLIP~\citep{dallago2021flip}, TAPE~\citep{rao2019evaluating}.  Table~\ref{table:method} compares \mname~with existing AI-based protein learning benchmarks. We extend the scope of existing protein learning benchmarks by incorporating more protein learning datasets, more cutting-edge deep learning models, and enhancing user-friendliness.

\begin{figure*}
\centering
\includegraphics[width=\linewidth]{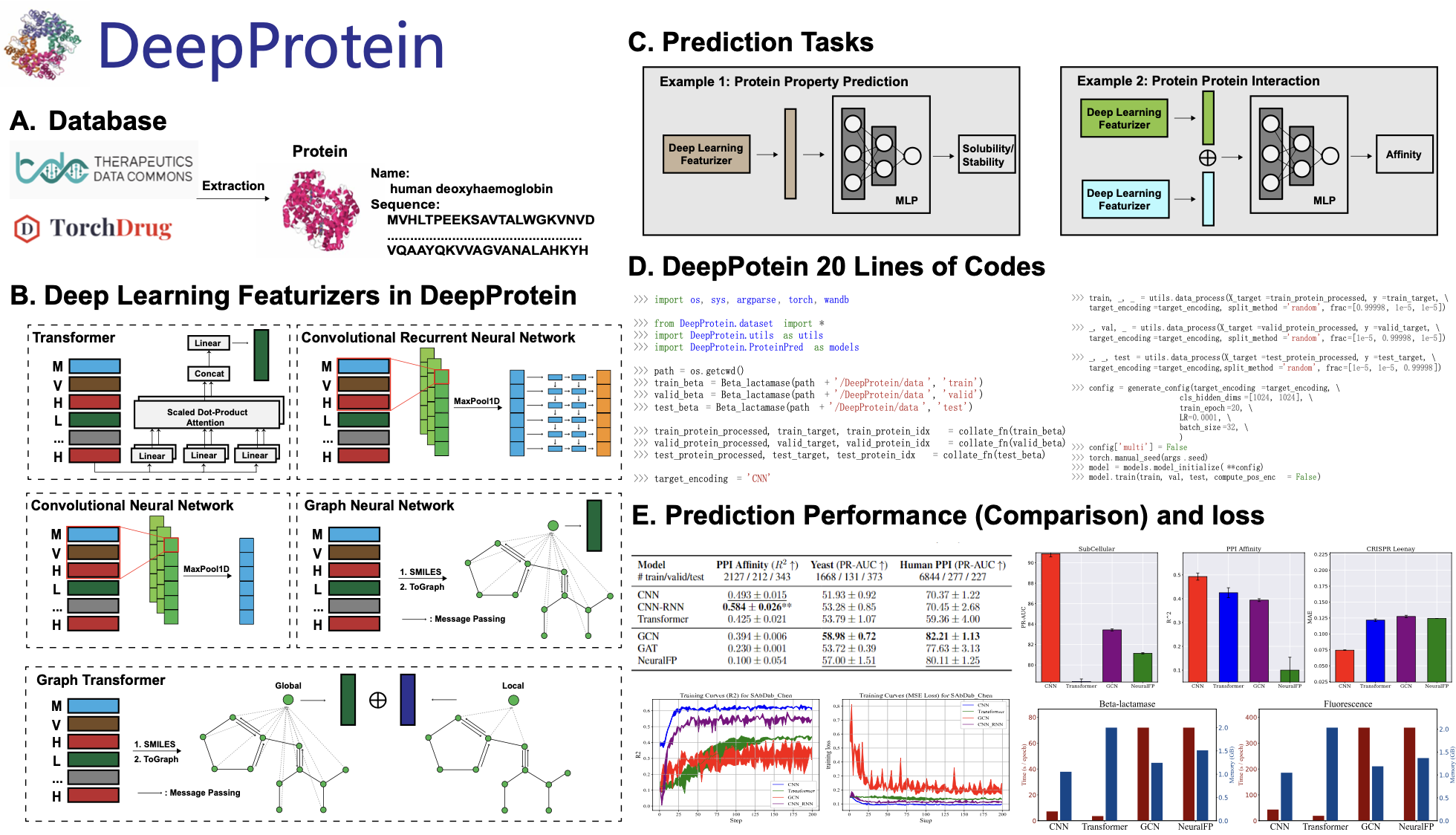}
\caption{\mname~framework. Part 1. A. \mname~is mainly selected from TorchDrug and Therapeutics Data Commons (TDC), where only protein tasks are considered, and all drug-related tasks are excluded, such as drug-target interactions. Specifically, DeepPurpose has established such a pipeline in their library. B. Both sequence-based and structure-based methods are included in \mname. For some graph neural networks, we utilized edge featurizers to generate additional edge information since inputs are 2-dimensional. Protein language models, large language models, and our pre-trained T5 (DeepProt-T5) are discussed in Figure 2. C. Task types are: protein function prediction, subcellular localization prediction, protein-protein interaction prediction, and protein structure prediction. They can be classified as either a 1 (protein)-to-1 (aim) problem or a 2 (proteins)-to-1 (aim) problem, which meets the researchers' needs. D. An earlier version of DeepProtein could be executed within 20 lines of code. The newest version of DeepProtein could be executed within 10 lines of code, where we further wrapped the data processing steps. Researchers can also provide their own data to either train or perform inference with the help of \mname. E. In this paper, we provide comprehensive results, including the performance of each model on corresponding tasks, the differences among sequence-based models, structure-based models, and pre-trained protein language models, and the computation resources, including time-stamps and GPU memory assumptions. As \mname~supports wandb, we also provide two wandb repositories that record the results of all experiments, which are \href{https://wandb.ai/jiaqing/DeepProtein?nw=nwuserjiaqing}{https://wandb.ai/jiaqing/DeepProtein?nw=nwuserjiaqing} and \href{https://wandb.ai/jiaqing/DeepPurposePP}{https://wandb.ai/jiaqing/DeepPurposePP}. Tables and figures are presented later in this paper.}
\label{fig:main}
\end{figure*}

\section{\mname~Library and Benchmark}\label{sec:main}

\subsection{AI-solvable Protein Problems}
In this section, we elaborate on a couple of AI-solvable protein problems and the related datasets. 
\begin{itemize}
\item \textbf{Protein Function Prediction}. Protein function prediction involves determining the biological roles and activities of proteins based on their sequences or structures. This process is crucial for understanding cellular mechanisms and interactions, as a protein's function is often linked to its sequence composition and the context of its cellular environment. Machine learning algorithms are employed to analyze known protein databases, identifying patterns and features that correlate with specific functions. Accurate predictions can facilitate drug discovery, help elucidate disease mechanisms, and support advancements in synthetic biology by providing insights into how proteins can be engineered for desired activities~\citep{zhang2021ddn2}. 
We consider the following datasets. 
\begin{itemize}
\item \textbf{Fluorescence}~\citep{sarkisyan2016local}. Protein fluorescence refers to the phenomenon where certain proteins can emit light of a specific wavelength when excited by light of a shorter wavelength. It is a widely used technique to study protein structure, dynamics, interactions, and function. The dataset consists of 54,025 protein sequences with real-valued groundtruth. The label is the logarithm of fluorescence intensity. 
\item \textbf{Stability}~\citep{rocklin2017global}. Protein stability is the capacity of a protein to preserve its three-dimensional structure and functional characteristics across different environmental conditions. This stability is essential for the proper functioning and longevity of proteins within biological systems. A protein's stability is influenced by its ability to withstand denaturation, aggregation, and degradation. The dataset comprises 68,934 protein sequences with real-valued groundtruth.   
\item $\beta$-\textbf{lactamase}~\citep{gray2018quantitative}. This task aims to predict the increased activity of $\beta$-lactamase, the most common enzyme that provides gram-negative bacteria with resistance to beta-lactam antibiotics through single mutations. The dataset consists of 5,198 protein sequences with real-valued groundtruth. The groundtruth refers to the experimentally determined fitness score, which measures the scaled mutation effect for each mutant. 
\item \textbf{Solubility}~\citep{khurana2018deepsol}. Protein solubility is the capacity of a protein to dissolve or remain dispersed in a solution. This property is crucial for determining how the protein behaves and functions in various biological and industrial contexts. Several factors influence a protein's solubility, including its amino acid composition, ionic strength, pH, temperature, and the presence of other molecules in the solution. The dataset consists of 71,419 protein sequences with binary labels.  
\end{itemize}
\item \textbf{Protein Localization Prediction}. Accurate localization predictions can enhance drug development by informing target identification and improving therapeutic efficacy, particularly in treating diseases linked to protein mislocalization. Additionally, insights gained from localization predictions facilitate the mapping of biological pathways, aiding in the identification of new therapeutic targets and potential disease mechanisms. 
\begin{itemize}
    \item \textbf{Subcellular}~\citep{almagro2017deeploc}. The task predicts the location of a natural protein within the cell. The dataset consists of 13,961 data samples with categorical labels (10 classes, $\{0,1,2,\cdots,9\}$).  
    \item \textbf{Binary}~\citep{almagro2017deeploc}. It is a simpler version of the previous task (10-category classification), where the model is trained to roughly forecast each protein as either ``membrane-bound'' or ``soluble'' (i.e., binary classification). The dataset comprises 8,634 data samples with binary labels. 
\end{itemize}
\item \textbf{Protein-Protein Interaction (PPI). } Proteins are the essential functional units in human biology, but they seldom operate in isolation; rather, they typically interact with one another to perform various functions. Understanding protein-protein interactions (PPIs) is crucial for identifying potential therapeutic targets for disease treatment. Traditionally, determining PPI activity requires costly and time-consuming wet-lab experiments. PPI prediction seeks to forecast the activity of these interactions based on the amino acid sequences of paired proteins. 
\begin{itemize}
  \item \textbf{PPI Affinity}~\citep{moal2012skempi}. It consists of 2,682 protein-protein pairs with real-valued groundtruth.  
  \item \textbf{Yeast}~\citep{guo2008using}. The dataset comprises 2,172 protein-protein pairs with binary labels.  
  \item \textbf{Human PPI}~\citep{pan2010large}. The dataset comprises 7,348 protein-protein pairs with binary labels.  
\end{itemize}
\item \textbf{Epitope Prediction}. 
An epitope, also known as an antigenic determinant, is the region of a pathogen that can be recognized by antibodies and cause an adaptive immune response. The epitope prediction task is to distinguish the active and non-active sites from the antigen protein sequences. Identifying the potential epitope is of primary importance in many clinical and biotechnologies, such as vaccine design and antibody development, and for our general understanding of the immune system~\citep{du2023abds}. In epitope prediction, the machine learning model makes a binary prediction for each amino acid residue. This is also known as \textbf{\textit{residue-level classification}}. 
\begin{itemize}
    \item \textbf{Immune Epitope Database (IEDB)}~\citep{vita2019immune}. It consists of 3,159 antigens with binary labels on each amino acid. The label indicates whether the amino acid belongs to the epitope, i.e., active position in binding. 
    It can be downloaded from TDC (\url{https://tdcommons.ai/single_pred_tasks/epitope/}).  
    \item \textbf{PDB-Jespersen}~\citep{jespersen2017bepipred}. It consists of 447 antigens with binary labels on each amino acid. It is curated by \citep{jespersen2017bepipred} and is extracted from PDB (Protein Data Bank). 
    It can be downloaded from TDC (\url{https://tdcommons.ai/single_pred_tasks/epitope/}). 
\end{itemize}
\item \textbf{Paratope Prediction}. Antibodies, or immunoglobulins, are large, Y-shaped proteins that can recognize and neutralize specific molecules on pathogens, known as antigens. They are crucial components of the immune system and serve as valuable tools in research and diagnostics. The paratope, also referred to as the antigen-binding site, is the region that specifically binds to the epitope. While we have a general understanding of the hypervariable regions responsible for this binding, accurately identifying the specific amino acids involved remains a challenge. This task focuses on predicting which amino acids occupy the active positions of the antibody that interact with the antigen. In paratope prediction, the machine learning model makes a binary prediction for each amino acid residue. This is also known as \textbf{\textit{residue-level classification}}. 
\begin{itemize}
    \item \textbf{SAbDab-Liberis}~\citep{liberis2018parapred} is curated from SAbDab~\citep{dunbar2014SAbDab}. It consists of 1,023 antibody chain sequences; each antibody contains both heavy and light chain sequences. It can be downloaded from TDC (\url{https://tdcommons.ai/single_pred_tasks/paratope/#sabdab-liberis-et-al}).  
\end{itemize}
\item \textbf{Antibody Developability Prediction}. Immunogenicity, instability, self-association, high viscosity, polyspecificity, and poor expression can hinder an antibody from being developed as a therapeutic agent, making early identification of these issues crucial. The goal of antibody developability prediction is to predict an antibody's developability from its amino acid sequences. A fast and reliable developability predictor can streamline antibody development by minimizing the need for wet lab experiments, alerting chemists to potential efficacy and safety concerns, and guiding necessary modifications. While previous methods have used 3D structures to create accurate developability indices, acquiring 3D information is costly. Therefore, a machine learning approach that calculates developability based solely on sequence data is highly advantageous. 
\begin{itemize}
\item \textbf{TAP}~\citep{raybould2019five}. It contains 242 antibodies with real-valued groundtruth.  Given the sequences of the antibody's heavy and light chains, we need to predict its developability (continuous value). The input consists of a list containing two sequences: the first representing the heavy chain and the second representing the light chain. It can be downloaded from TDC (\url{https://tdcommons.ai/single_pred_tasks/develop/}).  
\item \textbf{SAbDab-Chen}~\citep{chen2020predicting}. It consists of 2,409 antibodies with real-valued groundtruth. It is extracted from SAbDab (the structural antibody database)\footnote{It is publicly available \url{http://opig.stats.ox.ac.uk/webapps/newSAbDab/SAbDab/}. }, which is a database containing all the antibody structures available in the PDB (Protein Data Bank), annotated and presented in a consistent fashion~\citep{dunbar2014SAbDab}. Given the antibody's heavy chain and light chain sequence, predict its developability (binary label). It can be downloaded from TDC (\url{https://tdcommons.ai/single_pred_tasks/develop/}).  
\end{itemize}
\item \textbf{CRISPR Repair Outcome Prediction}. CRISPR-Cas9 is a gene editing technology that allows for the precise deletion or modification of specific DNA regions within an organism. It operates by utilizing a custom-designed guide RNA that binds to a target site upstream, which results in a double-stranded DNA break facilitated by the Cas9 enzyme. The cell responds by activating DNA repair mechanisms, such as non-homologous end joining, leading to a range of gene insertion or deletion mutations (indels) of varying lengths and frequencies. This task aims to predict the outcomes of these repair processes based on the DNA sequence. Gene editing marks a significant advancement in the treatment of challenging diseases that conventional therapies struggle to address, as demonstrated by the FDA's recent approval of gene-edited T-cells for the treatment of acute lymphoblastic leukemia. Since many human genetic variants linked to diseases arise from insertions and deletions, accurately predicting gene editing outcomes is essential for ensuring treatment effectiveness and reducing the risk of unintended pathogenic mutations. 
\begin{itemize}
    \item \textbf{CRISPR-Leenay}~\citep{leenay2019large}. The dataset comprises 1,521 DNA sequences (including guide RNA and PAM) with five measured repair outcomes, assessed across various donor populations of primary T cells. It can be downloaded from TDC (\url{https://tdcommons.ai/single_pred_tasks/CRISPROutcome/}). 
\end{itemize}
\end{itemize}

\begin{itemize}

  \item {\textbf{Protein Structure Prediction.} Protein structure prediction (PSP) is a fundamental problem in computational biology, aiming to determine the three-dimensional structure of a protein from its amino acid sequence. Specifically, in our benchmark, the task is to predict the family of a folding or secondary structure family that it belongs to. Since a protein’s structure dictates its function, accurate prediction is crucial for understanding the topology of the protein. PSP can be broadly divided into global topology prediction and local structural prediction, which include tasks such as fold classification and secondary structure prediction:}

  \begin{itemize}

    \item {\textbf{Fold} \citep{hou2018deepsf}. This task involves predicting the global structural topology of a protein, categorizing it into one of the predefined fold classes (within 0, 1, ..., 1194). During inference, we predict the class of each protein's superfamily. It contains 13766 samples overall.}

    \item {\textbf{Secondary Structure} \citep{klausen2019netsurfp}. This task focuses on predicting the local structural elements (coil, strand, helix) of each residue in a protein sequence. It serves as an intermediate step for more complex structure prediction tasks and is useful in applications such as functional analysis and multiple sequence alignment. This is a residue-level 3-class classification problem where the number of samples is equal to 11361.  }

  \end{itemize}

\end{itemize}

In this library, we follow the train-validation-test split in PEER benchmark \citep{xu2022peer} and TDC~\citep{huang2022artificial}. Each individual split is reported from Table \ref{tab:function} to \ref{tab:fold}.

\begin{figure*}
\centering
\includegraphics[width=\linewidth]{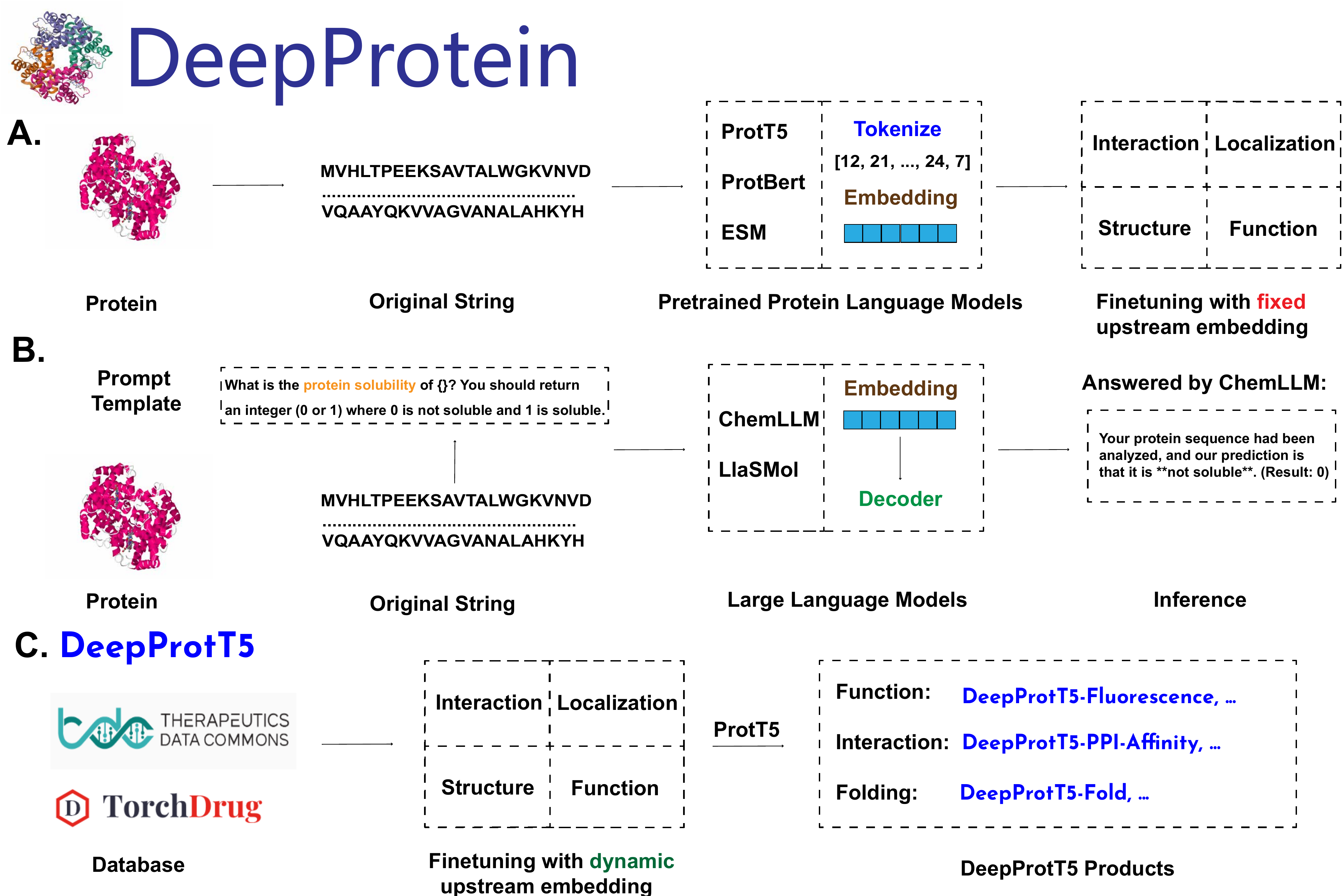}
\caption{\mname~framework, and DeepProt-T5. Part 2. A. For pre-trained protein language models, we directly use the initial string as the input instead of the transformed SMILES , since most of the protein language models have learned such representations. Strings are tokenized and carried as inputs to the model. We fine-tuned the models on the downstream tasks with the \textbf{fixed} embeddings, which means that after feature extraction, the upstream model parameters will not be trained any more. B. For large language models, such as ChemLLM and LlaSMol, we did not fine-tune them on the downstream tasks due to limited GPU resources. Instead we directly performed inference on downstream tasks where appropriate prompt templates should be carefully designed here. C. We extend the family of Prot-T5 to our DeepProt-T5, where the upstream embeddings are dynamic, by using the huggingface trainer. All fine-tuned models could be found here:\href{https://huggingface.co/jiaxie}{https://huggingface.co/jiaxie}.}
\label{fig:main_part2}
\end{figure*}

\subsection{Cutting-edge Deep Learning Methods}

At the core of deep learning lies the artificial neural network, a machine learning technique inspired by the architecture and functionality of the human brain. What distinguishes deep learning from other machine learning approaches is its exceptional ability to recognize and analyze complex, non-linear patterns in data, leading to enhanced performance and accuracy. Concretely, we incorporate several cutting-edge neural network architectures into two groups: 1) sequential-based learning and 2) structural-based learning. Detailed model architectures are described as follows:

\noindent\textbf{Sequential based learning} It generally takes a sequence as an input and uses one-hot encoding to pre-encode the input characters. Such learning methods include convolutional neural networks, recurrent neural networks, and transformers.
\begin{itemize}
\item \textbf{Convolutional Neural Network (CNN) (One-dimensional)} captures the local patterns in the data features, commonly used to analyze images and text. \textbf{(One-dimensional) Convolutional neural network (CNN)} takes amino acid sequences as the input. CNN has four layers; the number of filters for the four layers is 32, 64, and 96, respectively. The kernel sizes are 4, 8, and 12, respectively. The convolutional layer is followed by a one-layer MLP (multi-layer perceptron) to predict as a scalar.
\item \textbf{Recurrent Neural Network (RNN)} models sequence data and captures the long-term dependencies in the sequence data. RNN has two well-known variants: long short-term memory networks (LSTMs)~\citep{hochreiter1996lstm} and gated recurrent units (GRU)~\citep{Cho2014-ur}. The difference between GRU and LSTM is that GRU simplifies LSTM by removing the cell state and reducing the number of gates. We use a two-layer bi-directional GRU following three-layer CNN as the neural network architecture. The dimension of the hidden state in GRU is set to 64. ReLU function is applied after each GRU or CNN layer. 

\item \textbf{Transformer}~\citep{vaswani2017attention} architecture leverages the power of self-attention mechanisms and parallel computation to enhance the neural network's capability and efficiency in handling sequence data. We use the transformer encoder to represent the amino acid sequence. Two layers of transformer architectures are stacked. The dimension of embedding in the transformer is set to 64. The number of attention heads is set to 4. 
The ReLU function is applied after each self-attention layer. LayerNorm is applied after MLP layers.

\noindent\textbf{Structural-based learning} It generally transforms the input sequence into a valid SMILES string, then transforms the chemical substance into a graph. Then, graph filters are learned toward the input graph signal. Such learning methods are widely called Graph Neural Networks. Recently, graph transformers have shown their power in protein function prediction, and we included them as a part of structural-based learning.
\item \textbf{Graph Neural Network (GNN)} is a neural network architecture designed to process graph-structured data that takes input from nodes and edges, facilitating the flow of information between connected components to capture their interactions. It learns vector representations for both individual graph nodes and the overall graph structure. We consider the following GNN variants: 
\begin{itemize}
\item \textbf{Graph Convolutional Network (GCN)}~\citep{kipf2016semi}. GCN is a GNN variant that iteratively updates the node representation by aggregating the information from its neighbors. GCN has three layers, and the node embedding dimension is set to 64. After GCN, all the node embeddings are aggregated with a readout function (Weighted Sum and Max) to get graph-level embedding, followed by a one-layer MLP to get the final prediction.  BatchNorm is applied after MLP layers.
\item \textbf{Graph Attention Network (GAT)}~\citep{velivckovic2017graph}. GAT employs an attention mechanism to introduce anisotropy into the neighborhood aggregation function. This network features a multi-headed architecture that enhances its learning capacity. The node embedding dimension is 64. Readout function is the same as the one deployed in GCN model.
\item \textbf{Message Passing Neural Network (MPNN)}~\citep{gilmer2017neural}. MPNN is a GNN variant that considers passing messages (and modeling interactions) between both edges and nodes based on their neighbors. Edge features are included necessarily compared with GCN and GAT. Readout function is Sum And Max. Node and edge embedding dimension is 64.  
\item \textbf{Neural Fingerprint (NeuralFP)}~\citep{neuralfp}. NeuralFP uses Graph convolutional network (GCN)~\citep{kipf2016semi} to learn a neural network-based molecular embedding (also known as molecular \textit{neural fingerprint}, or NeuralFP) from a large amount of molecule data without labels. The neural fingerprint is essentially a real-valued vector, also known as embedding. Then, the neural fingerprint is fixed and fed into a three-layer MLP to make the prediction. Node embedding dimension is 64. BatchNorm is applied after MLP layers.
\item \textbf{Attentive Fingerprint (AttentiveFP)}~\citep{attentivefp}. AttentiveFP is a variant of graph neural networks that is enhanced by the attention mechanism when evaluating node and edge embedding. The model consists of three AttentiveFP layers with individual readout function: AttentiveFP readout. Node and edge embedding dimension is 64.
\end{itemize}
\item \textbf{Graph Transformer}~\citep{yun2019graph} is a type of neural network architecture designed to process graph-structured data by leveraging self-attention mechanisms. They extend the principles of traditional transformers, enabling them to capture the relationships and interactions between nodes in a graph effectively. 
\begin{itemize}
    \item \textbf{Path-Augmented Graph Transformer} (\textbf{PAGTN})~\citep{chen2019path}. It used augmented path features to capture long-range ($>$1 hop) graph properties. The model consists of 5 PAGTN layers with LeakyReLU activation. Node embedding dimension is 64.
    \item \textbf{Graphormer}~\citep{ying2021transformers}. It utilized transformer on graphs with spatial, centrality, and edge encoding. For simplicity and scalability on large graphs, we only deployed one Graphormer layer with ReLU activation. Node embedding dimension is 64. LayerNorm is applied after MLP layers.
\end{itemize}

\item \textbf{Foundation Model}. A foundation model is a large-scale, pre-trained machine learning model trained on extensive and diverse datasets, typically using self-supervised or unsupervised learning techniques. These models learn generalizable features and patterns from data, allowing them to perform various downstream tasks with minimal task-specific fine-tuning.  
\begin{itemize}

    \item \textbf{ESM}. {The Evolutionary Scale Modeling (ESM) utilizes large-scale pretraining on vast protein sequence data to capture evolutionary relationships and functional patterns within proteins~\citep{lin2023evolutionary, rives2021biological}. It benefits from Masked Language Modeling (MLM) and the transformer architecture. In this paper, we consider two ESM variants with different model sizes: ESM-1b and ESM-2-650M. The latter incorporates Rotary Position Embedding (RoPE) within the ESM-1 framework. We evaluate both models, where the embedding size is 1280.}
    \item \textbf{Prot-T5-XL}. {First introduced in ProtTrans~\citep{elnaggar2021prottrans}, Prot-T5-XL-UniRef50 is based on the T5-3B model and was pre-trained on a large corpus of protein sequences using a self-supervised approach. A key difference from the original T5 model is the denoising objective: while the original T5-3B model used a span denoising objective, this model employs a BART-like MLM denoising objective. The masking probability follows the original T5 training, randomly masking 15\% of the amino acids in the input. The embedding dimension is 1024.}
    \item \textbf{ProtBert}. {In addition to Prot-T5, ProtTrans includes a model pre-trained on BERT. Bidirectional Encoder Representations from Transformers (BERT) is a transformer-based neural network architecture pre-trained on unlabeled sequence data~\citep{devlin2018bert}. A key difference between ProtBert and the original BERT is how sequences are treated as separate documents, eliminating the need for next sentence prediction. The masking strategy follows the original BERT training, where 15\% of the amino acids in the input are randomly masked. The embedding dimension is 1024.
    }
\end{itemize}

\item \textbf{Large Language Model}. {In this paper, we distinguish foundation models as a class of pre-trained protein language models, whereas we define large language models (LLMs) as decoder-only models designed to generate sequential responses regarding the properties of one or multiple proteins given their sequences and a specific prompt. Due to computational constraints, fine-tuning 7B-scale models is resource-intensive; therefore, we focus on evaluating their performance instead. We consider two Protein LLMs in our study: ChemLLM-7B and LlaSMol-Mistral-7B. Their generalization ability to protein-related tasks has not been explored before. Additionally we provide the chat and prompt template in the appendix.}

\begin{itemize}
    \item \textbf{ChemLLM-7B}. {The backbone of ChemLLM-7B~\citep{zhang2024chemllmchemicallargelanguage} is the InternLM2-Base-7B model. It was initially trained on the Multi-Corpus-1.7M dataset on Hugging Face and later fine-tuned using instruction-tuning methods on ChemData (7M) and Multi-Corpus (1.7M). ChemLLM-7B has demonstrated superior performance over GPT-4 in retrosynthesis and temperature prediction tasks.  The model provides a set of predefined instruction-following templates, which are used in our study and detailed in the appendix.}
    
    \item \textbf{LlaSMol-Mistral-7B}. {The backbone of LlaSMol-Mistral-7B~\citep{yu2024llasmoladvancinglargelanguage} is Mistral-7B. It was fine-tuned using SMolInstruct, a large-scale, high-quality dataset designed for instruction tuning. SMolInstruct comprises 14 selected chemistry-related tasks and over three million samples, providing a solid foundation for training and evaluating LLMs in the field of chemistry. Specifically, in our experiments, we wrap the input SMILES sequences with $\langle \text{PROTEIN} \rangle \langle \text{/PROTEIN} \rangle$ token pairs to adapt them for protein-related tasks. The template is detailed in the appendix.}
\end{itemize}
\end{itemize}

\begin{itemize}
    \item \textsc{DeepProt-T5}. {We have trained Prot-T5-XL on our benchmark datasets individually (Figure \ref{fig:main_part2}). Different from the fixed embeedings, dynamic embeddings enabled us to finetuned the upstream architectures while also maintain a good predictive power for downstream tasks. This leads to a series of DeepProt-T5 models.}   
\end{itemize}

\noindent\textbf{Training setup.} For all models, the maximal training epoch number is set to 100. We used Adam optimizer~\citep{kingma2014adam} for training, with a default learning rate of 0.0001 for sequence-based learning and 0.00001 for structural-based learning. The batch size is equal to 32. More detailed hyper-parameter setups are listed in Table~\ref{flu_param} in the appendix. {For DeepProt-T5 models, fine-tuning has used a more generalized learning rate of 0.00002 for all models, batch size is equal to 10 to avoid memory errors.}

\begin{figure*}[!htb]
\centering
\includegraphics[width=0.9\linewidth]{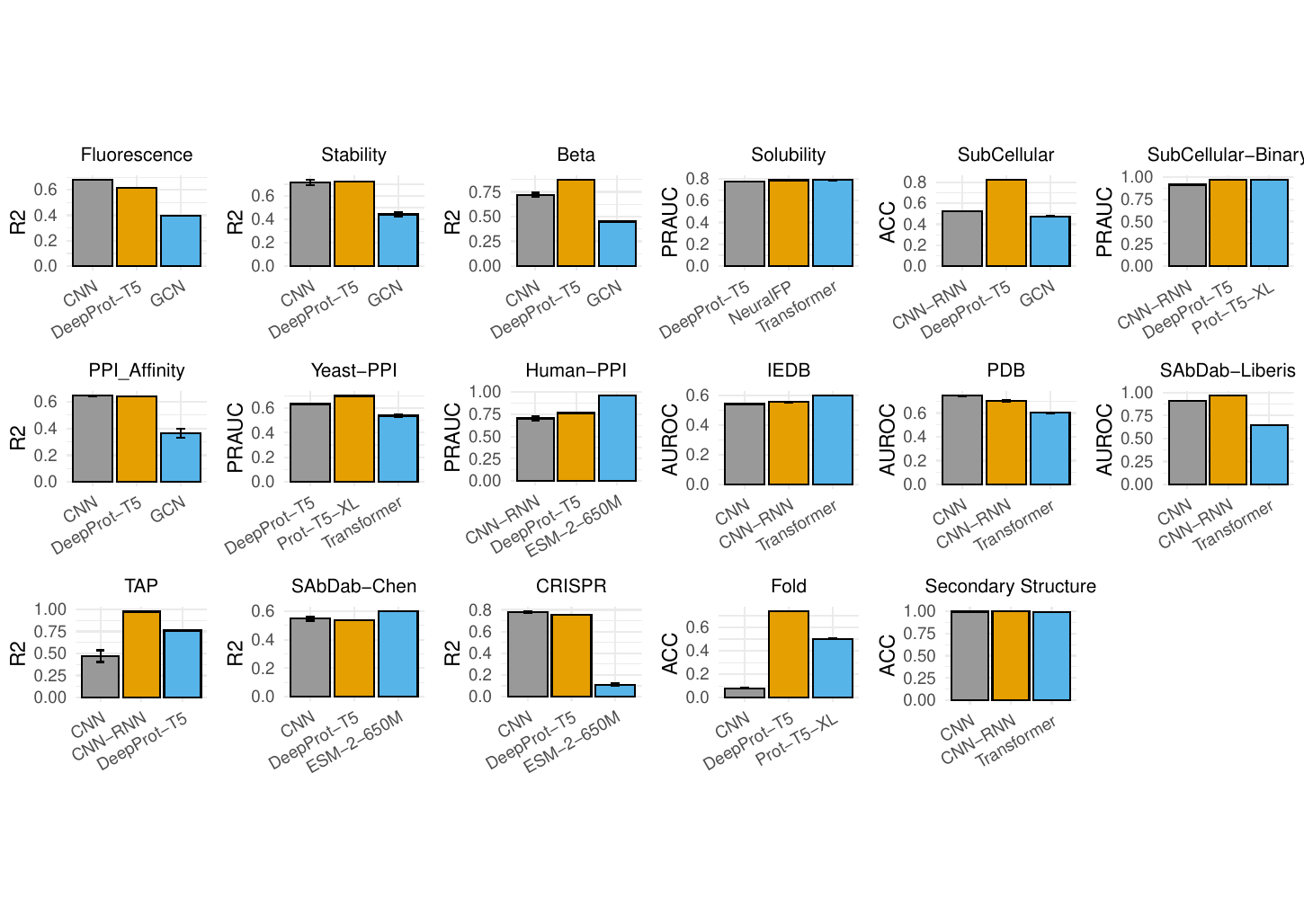}
\includegraphics[width=0.9\linewidth]{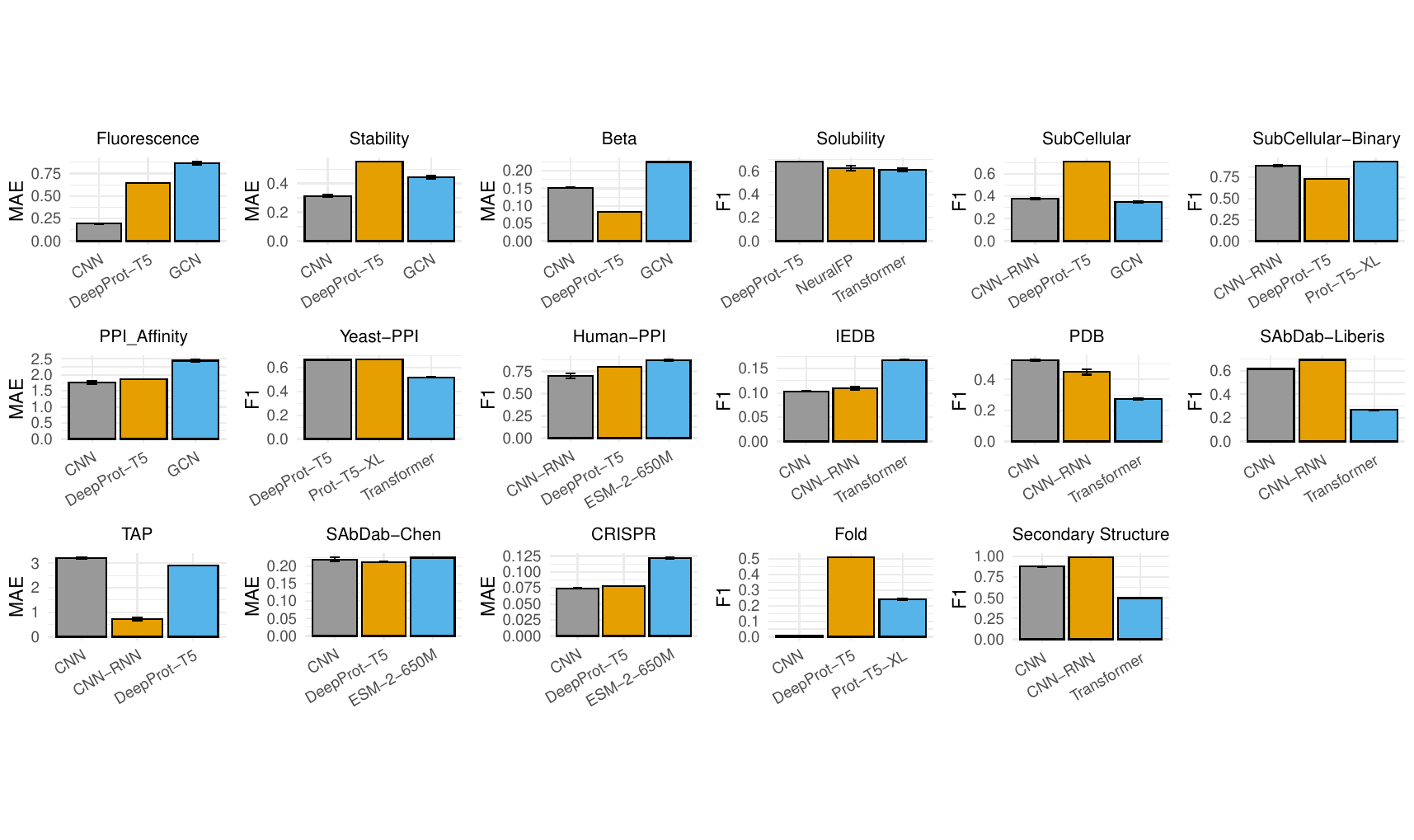}
\caption{Results of two metrics for selected deep learning methods for \mname~Benchmark. For regression task, metrics are Spearman (Pearson) Coefficient and Mean Absolute Error (MAE). For the binary classification task, metrics are ROC (or PR-AUC) and averaged macro F1. For multi classification task, metrics are the accuracy and averaged macro F1. Our DeepProt-T5 are competitive among deep learning methods included in our benchmark, and have improved original Prot-T5 models on six tasks: Beta-lactamase, Solubility, SubCellular, PPI\_Affinity, CRISPR and Fold.}
\label{fig:results_of_metrics}
\end{figure*}

\subsection{Experimental Setup and Implementation Details}

\noindent\textbf{Code Base.} This library is an extension of the well-established drug discovery library, DeepPurpose~\citep{huang2020deeppurpose}, building upon its foundational capabilities to offer enhanced features for protein-related tasks. By leveraging the strengths of DeepPurpose, this new library provides additional tools and functionalities tailored specifically for protein science. 
The library is publicly available at \url{https://github.com/jiaqingxie/DeepProtein/}. 

\noindent\textbf{Hardware Configuration.} All experiments that are mentioned in this paper were trained on a 40GB NVIDIA A40 and a 24GB NVIDIA RTX 3090. {For DeepProt-T5 fine-tuning, two 24GB NVIDIA RTX 3090 were used.} The parameters we provide have ensured the scalable training on these two types of GPUs. When running GNNs on protein localization tasks, we observed a large portion of GPU memory occupied irregularly, so we recommend cutting down the size of the number of workers from 8 to 4 or batch size from 32 to 8 or even smaller to potentially avoid GPU out-of-memory (OOM) problems. 

\noindent\textbf{Software Configuration.} The library is implemented in Python 3.9, PyTorch 2.3.0, PyTDC 0.4.1~\citep{huang2021therapeutics}, DeepPurpose 0.1.5~\citep{huang2020deeppurpose}, and RDKit 2023.9.6~\citep{landrum2006rdkit},  scikit-learn 1.2.2 \citep{pedregosa2011scikit}, and DGLlife 0.3.2 \citep{dgllife}. Besides, wandb is included in \mname~so that researchers can observe the visualization of training curves and test results easily. More details about environment setup could be found in the GitHub.

\begin{table*}[t!]
\centering
\caption{Results of protein function prediction. The $\uparrow$ symbol indicates that higher values are better for the corresponding metric. For each method, we employed five different random seeds to perform independent runs, reporting the average results along with their standard deviations. On each task, the best method is \textbf{bolded}, and the second (or the third closest to the second) best is \underline{underlined}. We use ``**'' to denote the method that achieves statistically better results than all the other methods (through statistical tests).}
\label{tab:function}
\resizebox{\linewidth}{!}{
\begin{tabular}{lcccccccc}
\toprule[1.0pt]
\textbf{Model} & \multicolumn{2}{c}{\textbf{Fluorescence}} & \multicolumn{2}{c}{\textbf{Stability}} & \multicolumn{2}{c}{$\beta$-\textbf{lactamase}} & \multicolumn{2}{c}{\textbf{Solubility}} \\ 
& $\rho$ $\uparrow$ & MAE $\downarrow$ & $\rho$ $\uparrow$ & MAE $\downarrow$ & $\rho$ $\uparrow$ & MAE $\downarrow$ & PR-AUC $\uparrow$& Averaged F1 $\uparrow$ \\
\# train/valid/test & \multicolumn{2}{c}{21446 / 5362 / 27217} & \multicolumn{2}{c}{53571 / 2512 / 12851} & \multicolumn{2}{c}{4158 / 520 / 520} & \multicolumn{2}{c}{62478 / 6942 / 1999} \\ 
\midrule
CNN & \textbf{0.680 $\pm$ 0.001} & \textbf{0.194 $\pm$ 0.001} &  \underline{0.715 $\pm$ 0.025}&  0.312 $\pm$ 0.008 &  \underline{0.721 $\pm$ 0.020} &   \underline{0.152 $\pm$ 0.001} & $74.61 \pm 0.55$ & 62.50 $\pm$ 2.69 \\
CNN-RNN & \underline{$0.678 \pm 0.001$} &  \underline{0.268 $\pm$ 0.002} & $0.635 \pm 0.025$ & 0.323 $\pm$ 0.001  & $0.695 \pm 0.012$ &  0.157 $\pm$ 0.002 &  $75.46 \pm 0.03$ &  60.87 $\pm$ 3.19\\
Transformer &  \underline{$0.648 \pm 0.001$} &  \underline{0.371 $\pm$ 0.005} & $0.375 \pm 0.052$ &  0.411 $\pm$ 0.026 & $0.310 \pm 0.041$ & 0.232 $\pm$ 0.007 & \bf 78.86 $\pm$ 0.46 & 61.28 $\pm$ 1.42 \\ 
\midrule
GCN & $0.397 \pm 0.002$ & 0.864 $\pm$ 0.018  & 0.443 $\pm$ 0.017 & 0.445 $\pm$ 0.011 & $0.450 \pm 0.006$ &0.225 $\pm$ 0.001  & $69.26 \pm 0.73$ & 56.30 $\pm$ 1.39  \\
GAT & $0.249 \pm 0.001$ & 1.201 $\pm$ 0.005 & $0.101 \pm 0.001$ & 0.740 $\pm$ 0.002  & $0.196 \pm 0.013$ & 0.264 $\pm$ 0.002 & $62.44 \pm 0.01$ & 47.79 $\pm$ 0.42 \\
NeuralFP & $0.349 \pm 0.049$ & 1.016 $\pm$ 0.081 & $0.373 \pm 0.023$ & 0.484 $\pm$ 0.020 & $0.171 \pm 0.007$ & 0.263 $\pm$ 0.001 &  \underline{$78.74 \pm 0.24$} &  \underline{62.72 $\pm$ 1.91}\\
AttentiveFP & $0.180 \pm 0.006$ &1.285 $\pm$ 0.001 & $0.013 \pm 0.004$ & 0.763 $\pm$ 0.003 & $0.058 \pm 0.011$ & 0.262 $\pm$ 0.007 & 60.56 $\pm$ 0.74 & 39.21 $\pm$ 0.66 \\
MPNN & $0.249 \pm 0.002$ & 1.275 $\pm$ 0.011 & $0.118 \pm 0.061$ & 0.732 $\pm$ 0.006 & $0.068 \pm 0.015$ & 0.262 $\pm$ 0.003 & $62.53 \pm 0.31$ & 40.56 $\pm$ 5.46\\ 
PAGTN & $0.139 \pm 0.014$ & 1.255 $\pm$ 0.040 & $0.088 \pm 0.194$ & 0.727 $\pm$ 0.011 & $0.092 \pm 0.018$ & 0.255 $\pm$ 0.007  & $61.33 \pm 0.91$ &  46.52 $\pm$ 4.13 \\
Graphormer & $0.058 \pm 0.015$ & 1.201 $\pm$ 0.032 & OOM & OOM & $0.067 \pm 0.046$ & 0.287 $\pm$ 0.007  & OOM & OOM \\ 
\midrule
{ESM-1b} & 0.562 $\pm$ 0.003& 0.548 $\pm$ 0.004& 0.682 $\pm$ 0.077 &  \underline{0.282 $\pm$ 0.010} &  0.664 $\pm$ 0.002& 0.171 $\pm$ 0.001 & Length Exceeded & Length Exceeded\\
{ESM-2-650M} & 0.548 $\pm$ 0.004 & 0.563 $\pm$ 0.002 &0.689 $\pm$ 0.020 &  \underline{0.273 $\pm$ 0.014} & 0.694 $\pm$ 0.001 & 0.154 $\pm$ 0.001 & 71.53 $\pm$ 0.58 & 63.33 $\pm$ 3.68 \\
{Prot-T5-XL} &  0.553 $\pm$ 0.003& 0.535 $\pm$ 0.006 &  $\bf 0.767 \pm 0.008$ & \bf 0.272 $\pm$ 0.030 &  \underline{ 0.756 $\pm$ 0.003} &  \underline{0.136 $\pm$ 0.003} &  74.95 $\pm$ 1.01 &  \underline{62.75 $\pm$ 0.76}  \\
{ProtBert} & 0.566 $\pm$ 0.001&0.583 $\pm$ 0.007 & 0.646 $\pm$ 0.064 & 0.322 $\pm$ 0.026& 0.572 $\pm$ 0.001 & 0.207 $\pm$ 0.001 & 70.70 $\pm$ 0.47 & 62.53 $\pm$ 2.67 \\
{\textsc{DeepProt-T5}} & 0.614& 0.642&  \underline{0.725} & 0.551 & \bf 0.874** & \bf 0.084** &  \underline{77.00} & \bf 68.25** \\
\midrule
{ChemLLM-7B} & -0.019 & 19.878 & -0.187& 46273.897 & 0.017& 111.76 & 50.65 & 1.64 \\
{LlaSMol-Mistral-7B} & NaN & 2.083 & 0.094 & 1.359 & NaN  & 0.268 & 50.30 & 6.89 \\
\bottomrule[1.0pt]
\end{tabular}
}
\end{table*}

\begin{table}[t!]
\centering
\caption{Results of protein localization prediction. 
}
\label{tab:localization}
\resizebox{0.65\linewidth}{!}{
\begin{tabular}{lcccc}
\toprule[1.0pt]

\textbf{Model} & \multicolumn{2}{c}{\textbf{Subcellular}} & \multicolumn{2}{c}{\textbf{Binary}}  \\
& Acc $\uparrow$ & Averaged F1 $\uparrow$ & PR-AUC $\uparrow$ & Averaged F1 $\uparrow$ \\
\# train/valid/test & \multicolumn{2}{c}{8,945 / 2248 / 2768} & \multicolumn{2}{c}{5161 / 1727 / 1746} \\ 
\midrule
CNN & 50.18 $\pm$ 1.21 & 30.44 $\pm$ 0.28 & $90.88 \pm 0.31$ & 87.74 $\pm$ 0.08\\
 CNN-RNN &  52.58 $\pm$ 0.11 & 38.21 $\pm$ 0.57 & 91.43 $\pm$ 0.45 &  88.56 $\pm$ 1.03\\
 Transformer & $42.63 \pm 0.68$ & 25.46 $\pm$ 0.01 & $78.38 \pm 0.25$ & 72.20 $\pm$ 2.21\\ \midrule
 GCN & $47.45 \pm 0.47$ & 34.88 $\pm$ 0.79 & $83.43 \pm 0.10$ & 81.80 $\pm$ 0.07\\
 GAT & $45.14 \pm 0.10$ & 27.03 $\pm$ 0.47 &$82.15 \pm 0.41$ & 81.89 $\pm$ 0.19\\
 NeuralFP & $45.20 \pm 0.49$ & 27.07 $\pm$ 0.93 &$81.14 \pm 0.06$ & 79.95 $\pm$ 0.09\\
 AttentiveFP  & $42.38 \pm 1.25$ & 23.50 $\pm$ 1.34 & $80.58 \pm 0.30$ & 80.08 $\pm$ 0.39 \\
\midrule
{ESM-2-650M} &  \underline{79.07 $\pm$ 0.05} &  \underline{66.66 $\pm$ 1.14} &  \underline{96.63 $\pm$ 0.18} &  \underline{91.76 $\pm$ 0.47} \\
{Prot-T5-XL} &  \underline{80.67 $\pm$ 0.04} &  \underline{69.86 $\pm$ 0.39} & \bf 97.03 $\pm$ 0.13 & \bf 93.48 $\pm$ 0.07\\
{LlaSMol-Mistral-7B} & 15.65 & 4.99 & 57.52 & 0.00  \\
{ChemLLM-7B} & 6.24 & 0.59 & 56.32 & 1.43\\
{\textsc{DeepProt-T5}} & \bf 82.69& \bf 82.52&  \underline{92.17} & \underline{92.18} \\
\bottomrule[1.0pt]
\end{tabular}
}
\end{table}

\begin{table}[t!]
\centering
\caption{Results of Protein-Protein Interaction (PPI). 
}
\label{tab:ppi}
\resizebox{0.78\linewidth}{!}{
\begin{tabular}{lccccccc}

\toprule[1.0pt]
\textbf{Model} & \multicolumn{2}{c}{\textbf{PPI Affinity}}  & \multicolumn{2}{c}{\textbf{Yeast PPI}}  &  \multicolumn{2}{c}{\textbf{Human PPI}}  \\
& $R^2$ $\uparrow$ & MAE $\downarrow$ & PR-AUC $\uparrow$ & Averaged F1 $\uparrow$  & PR-AUC $\uparrow$& Averaged F1 $\uparrow$ \\
\# train/valid/test & \multicolumn{2}{c}{2127 / 212 / 343} & \multicolumn{2}{c}{1668 / 131 / 373} & \multicolumn{2}{c}{6844 / 277 / 227} \\ 
 \midrule
 CNN & $\bf 0.646 \pm 0.003$ &\bf 1.764 $\pm$ 0.051 & $51.93 \pm 0.92$  & 25.90 $\pm$ 0.10 &$70.37 \pm 1.22$ &69.65 $\pm$ 1.63 \\
 CNN-RNN &   \underline{0.584 $\pm$ 0.026} &  \underline{1.886 $\pm$ 0.108} &  $53.28 \pm 0.85$ &  47.20 $\pm$ 3.62 &$70.45 \pm 2.68$ & 69.87 $\pm$ 2.59\\
 Transformer & $0.425 \pm 0.021$ & 2.081 $\pm$ 0.133 & $53.79 \pm 1.07$ &51.93 $\pm$ 0.40 & $59.36 \pm 4.00$ & 68.64 $\pm$ 1.06\\ \midrule
 GCN & $0.366\pm 0.034$ &  2.443 $\pm$ 0.036&  58.98 $\pm$ 0.72 & 48.13 $\pm$ 3.88 &   \underline{82.21 $\pm$ 1.13} & 71.05 $\pm$ 2.90\\
GAT & $0.230 \pm 0.001$ & 2.463 $\pm$ 0.015 &$53.72 \pm 0.39$ & 57.00 $\pm$ 3.83 &$77.63 \pm 3.13$ & 73.92 $\pm$ 3.50\\
NeuralFP & $0.100 \pm 0.054$ & 2.555 $\pm$ 0.040& $57.00 \pm 1.51$ & 58.94 $\pm$ 4.74& $80.11 \pm 1.25$ & 67.62 $\pm$ 1.03  \\
{ESM-2-650M} & 0.592 $\pm$ 0.001 & 1.893 $\pm$ 0.005 &  \underline{67.36 $\pm$ 0.80} &  \underline{63.99 $\pm$ 1.00} &  \bf 96.17 $\pm$ 0.18 & \bf 87.63 $\pm$ 0.60 \\
{Prot-T5-XL} & 0.573 $\pm$ 0.011 & 1.979 $\pm$ 0.007&  \bf 69.84 $\pm$ 0.46 & \bf 66.97 $\pm$ 0.01 &  \underline{95.36 $\pm$ 0.07} & \bf 87.64 $\pm$ 0.98\\
{LlaSMol-Mistral-7B} & -0.008 & 20.335 & 53.64 & 51.39 & 49.48 & 49.15\\
{ChemLLM-7B} & 0.082  & 28.771  &  53.65 & 15.45 & 47.15 & 15.49\\
{\textsc{DeepProt-T5}} &  \underline{0.643} &  \underline{1.870} &  \underline{63.20} &  \bf 66.70 & 76.37 & \underline{80.28}\\
    \bottomrule[1.0pt]
    \end{tabular}
}
\end{table}

\begin{table}[t!]
\centering
\caption{Results of epitope and paratope prediction  (\textit{residue-level classification}). Structure-based and pretrained protein language models took large GPU or CPU memory so we disregrad them in residue-level prediction. The same strategy is applied to secondary structure as well.}
\label{tab:epitope_paratope}
\resizebox{0.85\linewidth}{!}{
\begin{tabular}{lcccccc}

\toprule[1.0pt]
\textbf{Model} & \multicolumn{2}{c}{\textbf{IEDB}} & \multicolumn{2}{c}{\textbf{PDB-Jespersen}} &  \multicolumn{2}{c}{\textbf{SAbDab-Liberis}} \\
& ROC-AUC $\uparrow$ & Averaged F1 $\uparrow$ & ROC-AUC $\uparrow$ & Averaged F1 $\uparrow$ & ROC-AUC $\uparrow$ & Averaged F1 $\uparrow$ \\
\# train/valid/test & \multicolumn{2}{c}{2211 / 316 / 632} & \multicolumn{2}{c}{313 / 45 / 89} & \multicolumn{2}{c}{716 / 102 / 205}\\ 
\midrule
 CNN &  $54.03 \pm 0.02$ &  10.34 $\pm$ 0.02  & \bf 74.46 $\pm$ 0.21** & \bf 52.43 $\pm$ 0.35 & \underline{90.85 $\pm$ 0.08} &  \underline{61.51 $\pm$ 0.11} \\
 CNN-RNN & \underline{$55.47 \pm 0.23$} &  \underline{10.96 $\pm$ 0.26} & \underline{$70.10 \pm 0.97$} &  \underline{44.82 $\pm$ 1.69} & \bf 96.75 $\pm$ 0.10** & \bf 69.34 $\pm$ 0.10 \\
 Transformer & \bf 59.79 $\pm$ 0.06**  & \bf 16.79 $\pm$ 0.07 & $60.10 \pm 0.32$ & 27.31 $\pm$ 0.56&  $64.77 \pm 0.11$ & 26.73 $\pm$ 0.15\\ 

\bottomrule[1.0pt]
\end{tabular}
}
\end{table}

\begin{table}[t!]
\centering
\caption{Results of antibody developability prediction (TAP and SAbDab-Chen) and CRISPR repair outcome prediction (CRISPR-Leenay). }
\label{tab:developability_crispr}
\resizebox{0.8\linewidth}{!}{
\begin{tabular}{lcccccc}
\toprule[1.0pt]
\textbf{Model} &  \multicolumn{2}{c}{\textbf{TAP}} & \multicolumn{2}{c}{\textbf{SAbDab-Chen}}  &  \multicolumn{2}{c}{\textbf{CRISPR-Leenay}} \\
& $R^2$ $\uparrow$ & MAE $\downarrow$ & $R^2$ $\uparrow$ & MAE $\downarrow$  & $R^2$ $\uparrow$ & MAE $\downarrow$ \\
\# train/valid/test & \multicolumn{2}{c}{169 / 24 / 48} & \multicolumn{2}{c}{1686 / 241 / 482} & \multicolumn{2}{c}{1065 / 152 / 304} \\ 
\midrule
 CNN & 0.469 $\pm$ 0.066 &$3.217 \pm 0.026$ & \underline{0.547 $\pm$ 0.014} & 0.219 $\pm$ 0.006 & \bf 0.781 $\pm$ 0.009 &\bf 0.0745 $\pm$ 0.0005 \\
 CNN-RNN & \bf 0.972 $\pm$ 0.008** &\bf 0.712 $\pm$ 0.069** &0.486 $\pm$ 0.012 &\underline{$0.226 \pm 0.001$} & \underline{0.771 $\pm$ 0.004} &\underline{$0.0755 \pm 0.0010$} \\
 Transformer &0.030 $\pm$ 0.011 &$3.476 \pm 0.004$ & 0.452 $\pm$ 0.011 & $0.238 \pm 0.012$ & 0.200 $\pm$ 0.054 & $0.1216 \pm 0.0020$\\ \midrule
 GCN & 0.614 $\pm$ 0.054 & $2.761 \pm 0.155$& 0.434 $\pm$ 0.030 &$0.326 \pm 0.015$& 0.066 $\pm$ 0.010 &$0.1274 \pm 0.0019$\\
GAT & 0.777 $\pm$ 0.007 & $2.675 \pm 0.022$ & 0.356 $\pm$ 0.015 &$0.310 \pm 0.010$ &0.054 $\pm$ 0.010 &$0.1232 \pm 0.0001$ \\
NeuralFP & 0.205 $\pm$ 0.035& $3.436 \pm 0.015$ & 0.452 $\pm$ 0.027 &$0.253 \pm 0.011$ & 0.177 $\pm$ 0.029 & $0.1243 \pm 0.0001$\\
{ESM-2-650M} &  \underline{0.866 $\pm$ 0.007} & \underline{2.452 $\pm$ 0.027} & \bf 0.600 $\pm$ 0.001 & \underline{0.224 $\pm$ 0.001} & 0.110 $\pm$ 0.011 & 0.1218 $\pm$ 0.0010\\
{Prot-T5-XL} &  \underline{0.837 $\pm$ 0.010} & \underline{2.417 $\pm$ 0.056} & \underline{0.596 $\pm$ 0.011} & 0.229 $\pm$ 0.003 & 0.236 $\pm$ 0.004 & 0.1186 $\pm$ 0.0006\\
{LlaSMol-Mistral-7B} & NaN & 48.104 & NaN & 0.241 & NaN & 0.7922 \\
{ChemLLM-7B} & 0.099 &  45.957 & -0.006 & 32.33 & 0.061 & 20.0503 \\
{\textsc{DeepProt-T5}} & 0.758 & 2.922 & 0.536 & \bf0.186 & \underline{0.753} & \underline{0.078}\\
    \bottomrule[1.0pt]
    \end{tabular}
}
\end{table}

\begin{table}[t!]
\centering
\caption{ Results of protein folding prediction  (\textit{protein-level and residue-level classification}). }
\label{tab:fold}
\resizebox{0.7\linewidth}{!}{
\begin{tabular}{lcccc}

\toprule[1.0pt]
\textbf{Model} & \multicolumn{2}{c}{\textbf{Fold}} & \multicolumn{2}{c}{\textbf{Secondary Structure}}  \\
& Acc $\uparrow$ & Averaged F1 $\uparrow$ & Acc $\uparrow$ & Averaged F1 $\uparrow$  \\
\# train/valid/test & \multicolumn{2}{c}{12312 / 736 / 718 } & \multicolumn{2}{c}{ 8678 / 2170 / 513}\\ 
\midrule
 {CNN} & 8.01 $\pm$ 0.44 & 0.47 $\pm$ 0.11 & \underline{99.44 $\pm$ 0.00} & \underline{87.76 $\pm$ 0.06} \\
 {CNN-RNN} &  7.50  $\pm$ 0.01 & 1.06 $\pm$ 0.01 & \bf 99.93 $\pm$ 0.00 & \underline{98.80 $\pm$ 0.00} \\
 {Transformer} & 5.34 $\pm$ 0.24 &  0.20 $\pm$ 0.02 & 98.64 $\pm$ 0.00 & 49.66 $\pm$ 0.00\\ 
 \midrule
 {GCN} & 8.09 $\pm$ 0.36 & 2.34 $\pm$ 0.15 & / & /\\

 {ESM-2-650M} &  \underline{49.80 $\pm$ 0.36} &  \underline{25.55 $\pm$ 0.61}   & / & / \\
 {Prot-T5-XL} &  \underline{50.28 $\pm$ 0.52} &  \underline{24.24 $\pm$ 0.48}   & / & / \\
{LlaSMol-Mistral-7B} & 0.00 & 0.00 & / & / \\
{ChemLLM-7B} & 0.64  & 0.04  & / & / \\
{\textsc{DeepProt-T5}} & \bf 74.00** & \bf 73.61** & / & / \\

\bottomrule[1.0pt]
\end{tabular}
}
\end{table}

\subsection{Results \& Analysis}

For each method, we used three different random seeds to conduct independent runs and reported the average results and their standard deviations. 
The results of protein function prediction are reported in Table~\ref{tab:function} and Table~\ref{tab:localization}. 
The results of protein-protein interaction are reported in Table~\ref{tab:ppi}. 
The results of epitope and paratope interaction are reported in Table~\ref{tab:epitope_paratope}.  The results of antibody developability prediction are reported in Table~\ref{tab:developability_crispr}.  \textbf{The results of protein structure prediction are reported in Table~\ref{tab:fold}.}

\noindent\textbf{Statistical Test.} We also conduct statistical tests to confirm the superiority of the best-performed method compared with the second-best baseline method. The hypothesis is that the accuracies of the best method are the same as those of the baseline method. Student's T-test is used with significance level alpha as 1\% to calculate the p-values. When the p-values are below the 0.05 threshold, we reject the hypothesis and accept the alternative hypothesis, i.e., the best method is statistically significant compared with the second-best method. We use ``**'' to denote the method that achieves statistically better results than all the other methods (pass statistical tests).

\begin{figure}[!htb]
\centering
\includegraphics[width=1\linewidth]{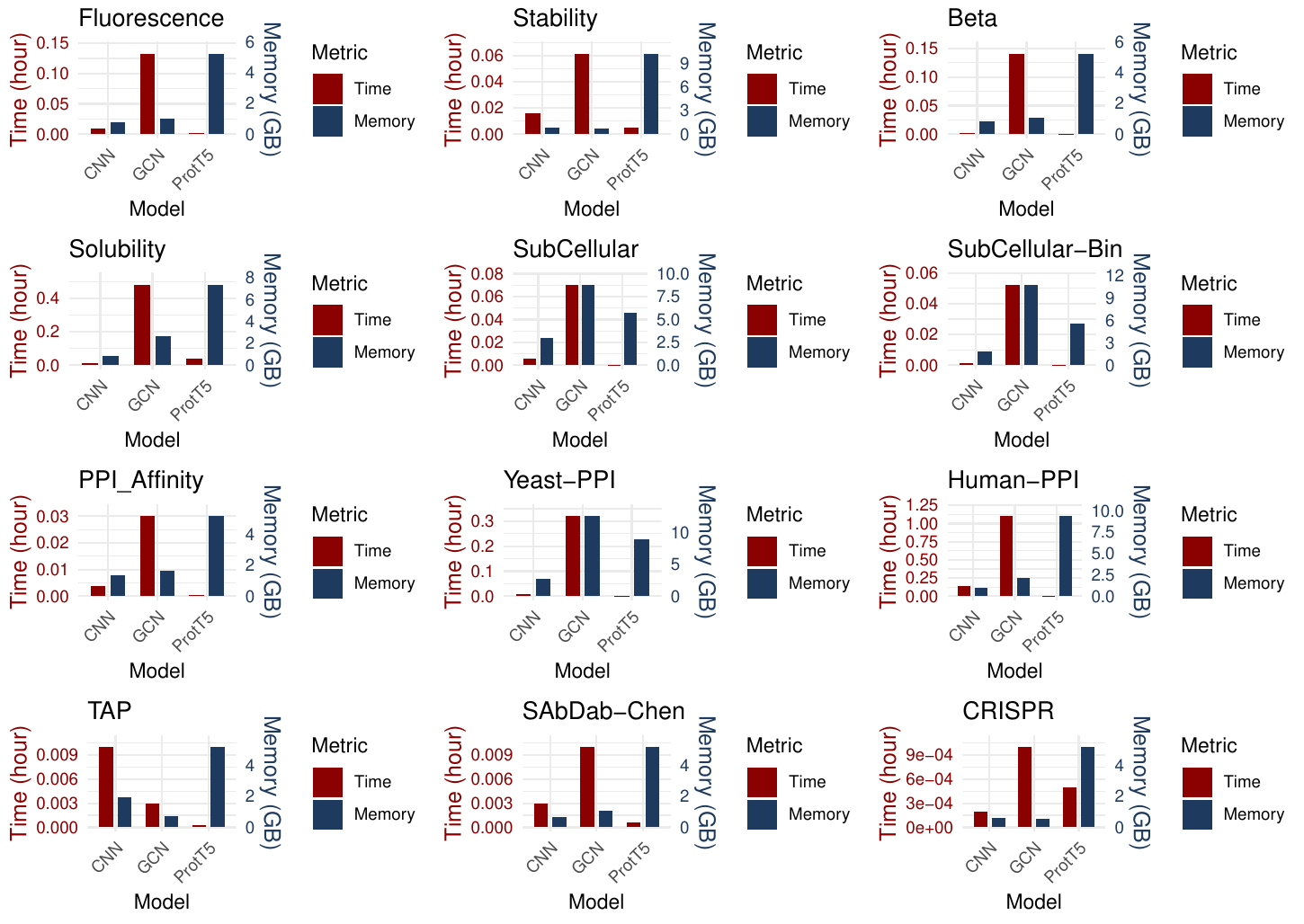}
\caption{We recorded training time and GPU memory assumptions for each task and each model. Specifically, we extract three representative methods: CNN, GCN and Prot-T5 from sequence-based, structure-based and pre-trained protein language models. We observed that Prot-T5 and GCN took up more GPU memory than CNN. Prot-T5, where upstream embeddings are fixed, is more efficient in training downstream tasks. Training a GCN model took more time than training a CNN or a Prot-T5 model. }
\label{fig:time_memory}
\end{figure}

\noindent\textbf{Key Observations.} We summarize the following key observations as takeaways. 
\begin{itemize}
\item {pre-trained protein language models and our DeepProt-T5 are powerful compared with sequence-based and structure-based neural architectures.} {
Sequence-based neural architectures, such as CNN, RNN, and transformer, obtain also superior performance in most protein sequence learning tasks. Specifically, in 12 out of all the 17 tasks across various protein sequence learning tasks, both sequenced-based models (CNN, RNN, Transformer) and the pre-trained protein language models (Prot-T5-XL, ESM-2-650M and DeepProt-T5) takes the top-2 position. }
\item{ Among all the 13 GNN-solvable tasks (except residue-level classification), graph neural networks (GNN) obtain the inferior performance compared with sequenced-based and protein language models. The potential reason would be that SMILES or original string didnt provide the 3d information (coordinates) about a protein, the graph topology given by edge featurizer is ill-defined in tbe deep graph library. }
\item Among all the graph neural networks (GNNs) across the whole 12 GNN-solvable tasks (except residue-level classification), the earliest variant, GCN~\citep{kipf2016semi}, achieves the best performance in 9 tasks. 
\item \textbf{Stability.} From the learning curve (Figure~\ref{fig:training_curve}), we find that GNN's training curve is not stable. In contrast, the sequence-based models, including CNN, RNN, and transformer, converge more stably from the learning curve. This can be observed from Figure~\ref{fig:training_curve}. On the contrary, training is more stable, fast and accurate for GAT when it comes to TAP dataset.
\item \noindent\textbf{Computational complexity}. The runtime and memory costs are reported in Figure~\ref{fig:time_memory}.  We find that GNN-based models are typically computationally inefficient. The key reason behind this is that GNN utilizes molecular graph as the feature, where each atom corresponds to a node and each chemical bond corresponds to an edge. While another model, such as CNN, RNN, and transformer, uses amino acid sequences as the input feature.  
\end{itemize}

\section{Conclusion} 
In this paper, we have developed \mname, which marks a significant advancement in the application of deep learning to protein science, providing researchers with a powerful and flexible tool to tackle various protein-related tasks. By integrating multiple state-of-the-art neural network architectures and offering a comprehensive benchmarking suite, \mname~empowers users to explore and optimize their models effectively. The detailed documentation and tutorials further enhance accessibility, promoting widespread adoption and reproducibility in research. As the field of proteomics continues to evolve, \mname~stands to contribute substantially to our understanding of protein functions, localization, and interactions, ultimately driving forward discoveries that can impact biotechnology and medicine. 



\appendix

\begin{figure}[!htb]
\centering
\includegraphics[width=0.8\linewidth]{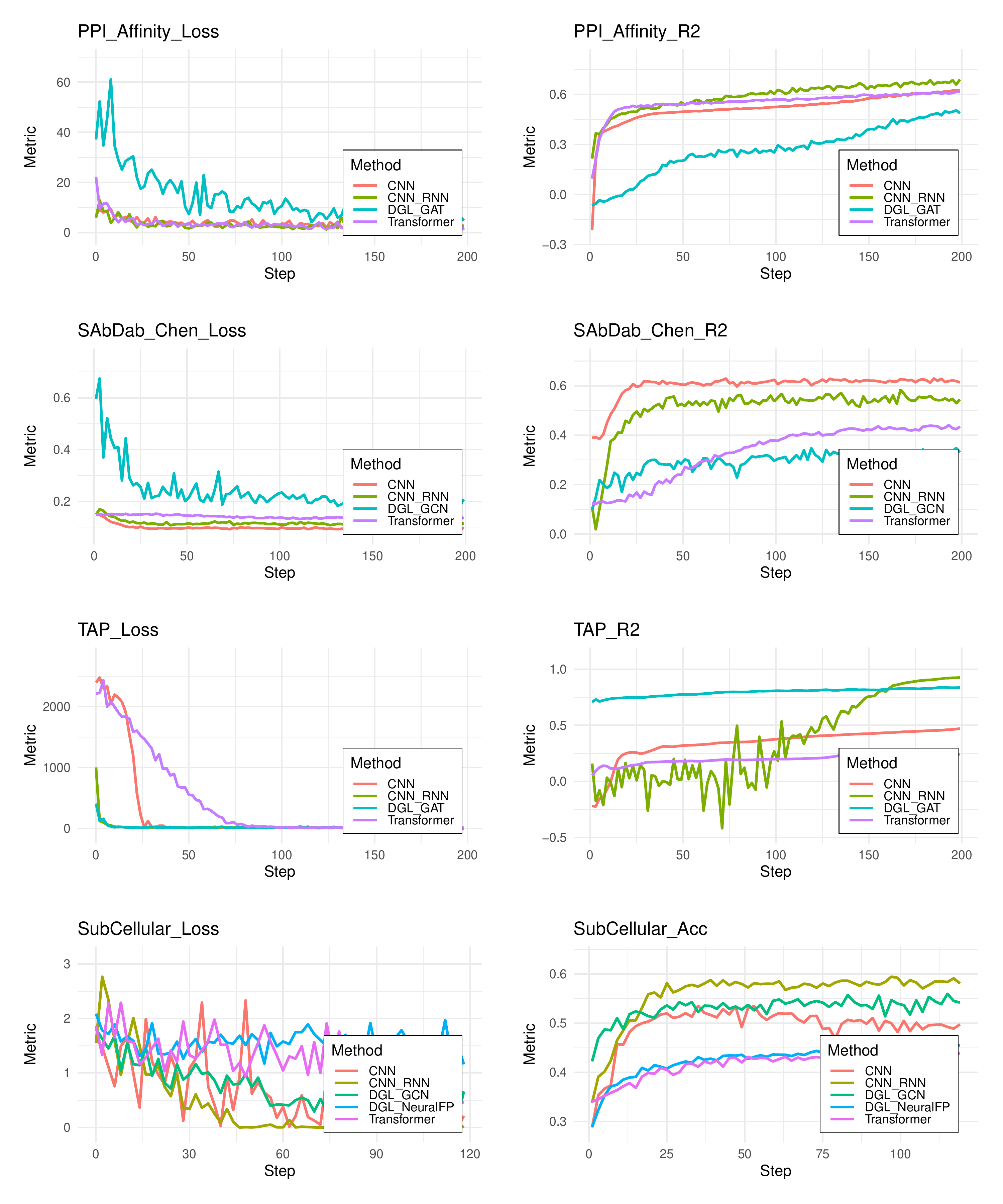}
\caption{Training Loss of selected datasets: PPI\_Affinity, SAbDab\_Chen, TAP, and SubCellular.}
\label{fig:training_curve}
\end{figure}

\section{Evaluation Metrics} 
In this section, we describe the basic evaluation metrics for both classification and regression tasks. {In the part optimization flow it would be further detailed on the end to end training flow.}

\noindent\textbf{Classification metrics.} 
Most classification tasks are binary classification, except subcellular prediction in protein localization prediction, which is a 10-category classification problem, where we use \textbf{accuracy (acc)} (the fraction of correctly predicted/classified samples) as the evaluation metric. In binary classification, there are four kinds of test data points based on their ground truth and the model's prediction, 
\begin{enumerate}
  \item positive sample and is correctly predicted as positive, known as \textit{True Positive (TP)};
  \item negative samples and is wrongly predicted as positive samples, known as \textit{False Positive (FP)};
  \item negative samples and is correctly predicted as negative samples, known as \textit{True Negative (TN)};
  \item positive samples and is wrongly predicted as negative samples, known as \textit{False Negative (FN)}. 
\end{enumerate}

\begin{itemize}
\item \textbf{Precision}. The precision is the performance of a classifier on the samples that are predicted as positive. 
It is formally defined as
$\text{precision} = \textit{TP} / (\textit{TP}+\textit{FP}). $

\item \textbf{Recall}. The recall score measures the performance of the classifier to find all the positive samples.
It is formally defined as 
$\text{recall} = \textit{TP} /(\textit{TP}+\textit{FN})$. 
\item \textbf{PR-AUC} (Precision-Recall Area Under Curve). The area under the Precision-Recall curve summarizes the trade-off between the true positive rate and the positive predictive value for a predictive model using different probability thresholds. 
\item \textbf{ROC-AUC} Area Under the Receiver Operating Characteristic Curve summarizes the trade-off between the true positive rate and the false positive rate for a predictive model using different probability thresholds. ROC-AUC is also known as the Area Under the Receiver Operating Characteristic curve (AUROC) in some literature. 
\end{itemize}
For all these metrics, the numerical values range from 0 to 1, a higher value represents better performance. 

\noindent\textbf{Regression metrics.} 
In the regression task, both ground truth and prediction are continuous values. 

\begin{itemize}
\item \textbf{Mean Squared Error (MSE)} measures the average of the squares of the difference between the forecasted value and the actual value. It is defined as 
$\text{MSE} = \frac{1}{N} \sum^{N}_{i=1} (y_i - {\hat{y}}_i)^2, $
where $N$ is the size of the test set; $y_i$ and $\hat{y}_i$ denote the ground truth and predicted score of the $i$-th data sample in the test set, respectively. 
MSE value ranges from 0 to positive infinity. 
A lower MSE value indicates better performance. 
\item \textbf{Mean Absolute Error (MAE)} measures the absolute value of the difference between the predicted value and the actual value. It is defined as 
$\text{MAE} = \frac{1}{N} \sum^{N}_{i=1} |y_i - {\hat{y}}_i|, $ where $N$ is the size of the test set; $y_i$ and $\hat{y}_i$ denote the ground truth and predicted score of the $i$-th data sample in the test set, respectively. 
MAE value ranges from 0 to positive infinity. 
It emphasizes the ranking order of the prediction instead of the absolute value. 
A lower MAE value indicates better performance. 
\item \textbf{Spearman rank correlation} ($\bf{\rho}$), also known as Spearman's $\rho$, is a nonparametric statistical test that measures the association between two ranked variables. A higher $\rho$ value indicates better performance. 
\item \textbf{R-squared ($R^2$) score} is defined as the proportion of the variation in the dependent variable that is predictable from the independent variable(s). It is also known as the coefficient of determination in statistics. Higher $R^2$ scores indicate better performance. 
\end{itemize}

\section{Optimization Flow}

\paragraph{Dataset Selection and Processing Flow} {As mentioned in the introduction part and Table \ref{table:method}, previous benchmarks either lack 1) the state-of-the-art deep learning methods 2) the diverse real-world data 3) easy-to-use files for researchers outside the computer science domain to use. Hence we collected the data from two main databases which contain approximately 20+ protein tasks which are enough for downstream testing. From them we deleted the tasks that were related to the drug, especially the task drug-target interaction since the DeepPurpose library supported such functionality. For the datasets in the PEER benchmark, \mname~just inherited the functions which transformed the data into standard torch datasets, and for the TDC data they were transformed to the standard torch dataset similarly. When loading the dataset, it will load a pair of (protein sequence, aim) or a triple of (protein sequence 1, protein sequence 2, aim), depending on the task type.}
\paragraph{Featurization Flow}

Since we are talking about training here instead of inference, we ignore the featurization flow of large language models. We mainly consider three types of methods here, which are sequence-based, structure-based and pretraind protein language models. 

\textbf{Sequence-based} models take the tokenized SMILES string as the input $\text{X}$, 
\begin{equation}
    \begin{aligned}
        \textbf{CNN:} &\quad X^{(l)} = X^{(l-1)} * W^{(l)} + b^{(l)} \\
        \textbf{RNN:} &\quad h_t = \sigma(W_h X_t + U_h h_{t-1} + b_h) \\
        \textbf{Attention:} &\quad \text{Attention}(Q, K, V) = \text{softmax} \left( \frac{Q K^T}{\sqrt{d_k}} \right) V
    \end{aligned}
\end{equation}
{In CNN, $W^{(l)}$ is the weight matrix at layer l which is convoluted by the last hidden layer input  $X^{(l-1)}$ and $b^{(l)}$ is the bias. The hidden state is decided by $X^{(l)}$. In RNN, we take each token in $X$ as the input at each time step t, where $W_h$ is the weight matrix for the current input token (amino acid) $X_t$ and $U_h$ is the weight matrix for the last hidden state and $b_h$ is the bias. We note that the protein sequence would be long in real-world data, so we truncate them to the maximum length 300, which also avoids memory exploding in RNN. In transformer, for each attention block, we could compute Q, K, V by $W_Q X$, $W_K X$, and $W_V X$, then attention is computed by equation (1). Noted that we could aggregated heads of attention to perform multi-head attention. }

\textbf{Structure-based} models take the graph $\mathcal{G}$ as the input, with node features $\mathbf{H}^{(0)}$ and adjacency matrix $\mathbf{A}$. Edge features could be added well if it's well prepared by the dataset. Especailly for the 2D protein structure, we could only obtain node features by using the features from CNN for instance. GCN, GAT and Graph Transformer's forms are given by:

\begin{equation}
\begin{aligned}
\textbf{GCN:} & \quad \mathbf{H}^{(l+1)} = \sigma \left( \tilde{\mathbf{D}}^{-\frac{1}{2}} \tilde{\mathbf{A}} \tilde{\mathbf{D}}^{-\frac{1}{2}} \mathbf{H}^{(l)} \mathbf{W}^{(l)} \right) \\
\textbf{GAT:} & \quad \mathbf{h}_i^{(l+1)} = \sigma \left( \sum_{j \in \mathcal{N}(i) \cup \{i\}} \alpha_{ij}^{(l)} \mathbf{W}^{(l)} \mathbf{h}_j^{(l)} \right), \\
& \quad \alpha_{ij}^{(l)} = \frac{\exp \left( \text{LeakyReLU} \left( \mathbf{a}^{\top} \left[ \mathbf{W}^{(l)} \mathbf{h}_i^{(l)} \| \mathbf{W}^{(l)} \mathbf{h}_j^{(l)} \right] \right) \right)}{\sum_{k \in \mathcal{N}(i) \cup \{i\}} \exp \left( \text{LeakyReLU} \left( \mathbf{a}^{\top} \left[ \mathbf{W}^{(l)} \mathbf{h}_i^{(l)} \| \mathbf{W}^{(l)} \mathbf{h}_k^{(l)} \right] \right) \right)} \\
\textbf{Graph Transformer:} & \quad \mathbf{H}^{(l+1)} = \text{Softmax} \left( \frac{\mathbf{Q} \mathbf{K}^\top}{\sqrt{d}} + \mathbf{A} \right) \mathbf{V}, \\
& \quad \mathbf{Q} = \mathbf{H}^{(l)} \mathbf{W}_Q, \quad \mathbf{K} = \mathbf{H}^{(l)} \mathbf{W}_K, \quad \mathbf{V} = \mathbf{H}^{(l)} \mathbf{W}_V \\
\textbf{MPNN:} & \quad \mathbf{m}_v^{(l)} = \sum_{u \in \mathcal{N}(v)} M\left(\mathbf{h}_v^{(l)}, \mathbf{h}_u^{(l)}, \mathbf{e}_{vu} \right), \\
& \quad \mathbf{h}_v^{(l+1)} = U\left(\mathbf{h}_v^{(l)}, \mathbf{m}_v^{(l)}\right)
\end{aligned}
\end{equation}
{In GCN, $\mathbf{H}^{(l)}$ is the node representation at layer l, $\tilde{\mathbf{A}}$ is the adjacency matrix with self loop added. $\tilde{\mathbf{D}}$ is the degree matrix corresponding to $\tilde{\mathbf{A}}$. $\mathbf{W}^{(l)}$ is the weight matrix at layer l. in GAT, $\alpha_{ij}^{(l)}$ works as a trainable attention parameter to present the attention betwen node i and node j at layer l. In a general graph transformer, adjacency matrix is added to the attention term which is different from the vanilla self-attention block. Therefore, the complexity is still $O(n^2)$ if there're $n^2$ nodes in the graph. In the message passing neural network (MPNN), additional edge information $e_{vu}$ is considered for node v and node u.}s

{For the pre-trained protein languade models (PLM), the general form could be written as:}
\begin{equation}
    \textbf{X}' = \textbf{PLM} (\textbf{X})
\end{equation}
{where we regard PLM as a white box model. We could get the embedding for the whole protein sequence instead of encoding each amino acid one by one which is more efficient than sequence-based or structure-based encoding.} 

\paragraph{Training Flow}

{After we obtained features from the featurizer module, we train the downstream tasks with a linear layer with the weight $\textbf{W}$ and bias $b$. We consider five machine learning task types (not referring to protein learning tasks), which are single protein regression, single protein classification, protein pair regression, protein pair classification, and token (residue) level single protein classification. We introduce them one by one.}

{Single protein regression task is that given a single protein's representation X, after applying the linear layer, we got a floating-point number $\hat{y}$, so mean squared error loss between the true value $y_i$ and predicted value $\hat{y}_i$ is applied during training. For single protein classification task, we apply the softmax function after the linear layer to decide its class. Either a binary cross-entropy loss (BCELoss) or a general cross-entropy loss (CELoss) would be back-propagated during the training:}
\begin{flalign*}
& \text{Single protein regression:} \quad \hat{y} = \mathbf{W} X + b && \\
& \mathcal{L}_{\text{MSE}} = \frac{1}{N} \sum_{i=1}^{N} (\hat{y}_i - y_i)^2 && \\
& \text{Single protein classification (multi-class):} \quad \hat{y} = \text{Softmax}(\mathbf{W} X + b) && \\
& \mathcal{L}_{\text{CE}} = -\sum_{i} y_i \log(\hat{y}_i) &&
\end{flalign*}
{Protein pair regression is that, the input is a pair of protein ($X_i, X_j$), the aim is to predict its affinity or some other related interaction metrics, labeled $y_{ij}$ here. The representation of two proteins is concatenated before it is applied to a linear layer. The predicted value is $\hat{y}_{ij}$. MSE loss between $\hat{y}_{ij}$ and $y_{ij}$ would be back-propagated. For protein pair classification task, we apply a sigmoid function as all labels are either 0 or 1 in our benchmark. BCELoss is being computed. }
\begin{flalign*}
& \text{Protein pair regression:} \quad \hat{y}_{ij} = \mathbf{W} (X_i \| X_j) + b && \\
& \mathcal{L}_{\text{MSE}} = \frac{1}{N} \sum_{i,j} (\hat{y}_{ij} - y_{ij})^2 && \\
& \text{Protein pair classification:} \quad \hat{y}_{ij} = \sigma(\mathbf{W} (X_i \| X_j) + b) && \\
& \mathcal{L}_{\text{BCE}} = -\sum_{i,j} y_{ij} \log(\hat{y}_{ij}) + (1 - y_{ij}) \log(1 - \hat{y}_{ij}) &&
\end{flalign*}
{For residue-level single protein classification, we predict the class for each token (amino acid) for each protein sequence, from the token $X_1$ to the the token $X_T$ if the length is equal to T. A softmax is applied here after applying a linear layer and CE-loss is calculated. Note that it is computation inefficient to perform residue level prediction when applying graph neural networks and protein language models and could easily reach memory bound so in our benchmark we only tested those datasets with CNN, CNN-RNN and Transformer architectures. $\hat{Y}_{t, c}$ is the probability that $X_t$ is being assigned to class c.}
\begin{flalign*}
& \text{Residue-level single protein classification:} \quad \hat{Y}_t = \text{Softmax}(\mathbf{W} X_t + b), \quad t = 1, \dots, T && \\
& \mathcal{L}_{\text{CE}} = -\frac{1}{T} \sum_{t=1}^{T} \sum_{c} Y_{t, c} \log(\hat{Y}_{t, c}) && \\
\\
\end{flalign*}

\section{Tables of Time and Memory Usage}

\begin{table}[!htb]
\centering
\caption{Memory and Time Usage of Different Models}
\label{tab:memory_time}
\begin{tabular}{llrr}
\toprule
        Dataset &  Model &  Time (hour) &  GPU Memory (GB) \\
\midrule
   Fluorescence &    CNN &      0.01000 &          0.78000 \\
   Fluorescence &    GCN &      0.13200 &          1.00000 \\
   Fluorescence & Prot-T5 &      0.00160 &          5.20000 \\
      Stability &    CNN &      0.01600 &          0.84480 \\
      Stability &    GCN &      0.06100 &          0.73656 \\
      Stability & Prot-T5 &      0.00472 &         10.19000 \\
           Beta &    CNN &      0.00150 &          0.84700 \\
           Beta &    GCN &      0.14000 &          1.07000 \\
           Beta & Prot-T5 &      0.00060 &          5.20000 \\
     Solubility &    CNN &      0.01300 &          0.84700 \\
     Solubility &    GCN &      0.48000 &          2.70200 \\
     Solubility & Prot-T5 &      0.04000 &          7.33440 \\
    SubCellular &    CNN &      0.00600 &          2.94500 \\
    SubCellular &    GCN &      0.07000 &          8.72600 \\
    SubCellular & Prot-T5 &      0.00060 &          5.71200 \\
SubCellular-Bin &    CNN &      0.00110 &          1.82600 \\
SubCellular-Bin &    GCN &      0.05200 &         10.75200 \\
SubCellular-Bin & Prot-T5 &      0.00030 &          5.62300 \\
   PPI\_Affinity &    CNN &      0.00400 &          1.37000 \\
   PPI\_Affinity &    GCN &      0.03000 &          1.64600 \\
   PPI\_Affinity & Prot-T5 &      0.00050 &          5.16200 \\
      Yeast-PPI &    CNN &      0.01000 &          2.69000 \\
      Yeast-PPI &    GCN &      0.32000 &         12.62000 \\
      Yeast-PPI & Prot-T5 &      0.00020 &          9.04000 \\
      Human-PPI &    CNN &      0.14000 &          0.96000 \\
      Human-PPI &    GCN &      1.10000 &          2.13000 \\
      Human-PPI & Prot-T5 &      0.00320 &          9.38000 \\
            TAP &    CNN &      0.01000 &          1.95000 \\
            TAP &    GCN &      0.00300 &          0.69100 \\
            TAP & Prot-T5 &      0.00030 &          5.16000 \\
    SAbDab-Chen &    CNN &      0.00300 &          0.65000 \\
    SAbDab-Chen &    GCN &      0.01000 &          1.05000 \\
    SAbDab-Chen & Prot-T5 &      0.00060 &          5.16000 \\
         CRISPR &    CNN &      0.00020 &          0.62900 \\
         CRISPR &    GCN &      0.00100 &          0.53800 \\
         CRISPR & Prot-T5 &      0.00050 &          5.16000 \\
           Fold &    CNN &      0.01000 &          0.92200 \\
           Fold &    GCN &      0.06000 &          1.80700 \\
           Fold & Prot-T5 &      0.00020 &          5.22200 \\
\bottomrule
\end{tabular}
\end{table}

\section{Prompt Template}
Template for ChemLLM-7B:
\begin{promptbox}
\textbf{System Instruction:} \\
\texttt{You are an AI assistant specializing in protein property prediction. Follow the given instruction format.}
\textbf{User Prompt Format:} \\
\texttt{
<|im\_start|>user\\
What is the \{protein\_property\} of the given protein sequence \{protein\_sequence\}? \{instruction\}\\
<|im\_end|>\\
<|im\_start|>assistant
}
\end{promptbox}
Template for LlaSMol-Mistral-7B:
\begin{promptbox}
What is the \{protein\_property\} of the given protein sequence $\langle$PROTEIN$\rangle$ \{protein\_sequence\} $\langle$\textbackslash PROTEIN$\rangle$? \{instruction\}
\end{promptbox}

Instruction and property (task) for each dataset.

\begin{longtable}{| m{3cm} | m{6cm} | m{5cm} |}
    \hline
   \textbf{Datasets} & \textbf{Instruction} & \textbf{Property} \\
    \hline
    Fluorescence & \RaggedRight You should return a floating-point number. & \RaggedRight Fluorescence intensity \\
    \hline
    Beta & \RaggedRight You should return a floating-point number. & \RaggedRight Increased activity \\
    \hline
    Stability & \RaggedRight You should return a floating-point number. & \RaggedRight Protein stability \\
    \hline
    Solubility & \RaggedRight You should return an integer (0 or 1) where 0 is not soluble and 1 is soluble. & \RaggedRight Protein solubility \\
    \hline
    Subcellular & \RaggedRight You should choose an integer within the range [0, 9] to indicate the protein's location. & \RaggedRight Location \\
    \hline
    Subcellular\_Binary & \RaggedRight You should return an integer (0 or 1) where 0 is membrane-bound and 1 is soluble. & \RaggedRight Location \\
    \hline
    Tap & \RaggedRight You should return a floating-point number. & \RaggedRight Developability \\
    \hline
    SAbDab\_Chen & \RaggedRight You should return a floating-point number. & \RaggedRight Developability \\
    \hline
    CRISPR & \RaggedRight You should return a floating-point number. & \RaggedRight Repair outcome \\
    \hline
    PPI-Affinity & \RaggedRight You should return a floating-point number. & \RaggedRight Activity of protein-protein interaction \\
    \hline
    Yeast-PPI & \RaggedRight You should return an integer (0 or 1) where 0 is weak and 1 is strong. & \RaggedRight Activity of protein-protein interaction \\
    \hline
    Human-PPI & \RaggedRight You should return an integer (0 or 1) where 0 is weak and 1 is strong. & \RaggedRight Activity of protein-protein interaction \\
    \hline
    Fold & \RaggedRight You should return an integer within the range [0, 1194]. & \RaggedRight Global structural topology of a protein on the fold level \\
    \hline
    Secondary & \RaggedRight You should return an integer within the range [0, 2]. & \RaggedRight Local structures of protein residues in their natural state \\
    \hline
\end{longtable}
\section{Hyperparameter Settings}

In table~\ref{flu_param}, we have listed a common settings of hyperparameter used in this library. In terms of learning rate (lr), a higher learning rate which is equal to 0.0001 for graph neural networks would lead to failure in training. For Subcellular and its binary version, a training epoch of 60 is enough for convergence. For small-scale protein datasets such as IEDB~\citep{yi2018enhance}, PDB-Jespersen, and SAbDab-Liberis, a larger learning rate of 0.001 also leads to convergence and the same performance when using CNN, CNN-RNN, and Transformer. For TAP, SAbDab-Chen and CRISPR-Leenay, larger learning rate of 0.0001 is suggested when training graph neural networks.

\begin{table}[!htb] 
\caption{Default Model Configurations for Protein Sequence Learning.}
\label{flu_param}
\centering
\resizebox{0.8167\textwidth}{!}{%
\begin{tabular}{lccccccccccc}
\toprule
Model & lr & dropout & activation & \# heads & \# layers & hidden dim & pooling & batch size & \# epochs & norm   \\
\midrule
CNN & $10^{-4}$ & 0.1 & ReLU & - & 3 & 256 & MaxPool1d & 32 & 100 & - \\ 
CNN-GRU & $10^{-4}$ & 0.1 & ReLU &  - & 2 & 64 & - & 32 & 100 & -   \\
Transformer & $10^{-4}$ & 0.1 &  ReLU & 4 & 2 & 64 & - & 32 & 100 & LayerNorm  \\
GCN & $10^{-5}$ & 0.1 & ReLU  &- & 3 & 64 & Weighted Sum + Max & 32 & 100 & BatchNorm   \\
GAT & $10^{-5}$ & 0.1 & ReLU &- & 3 & 64 & Weighted  Sum + Max & 32 & 100 & -  \\
NeuralFP & $10^{-5}$ & 0.1 & ReLU  &- & 3 & 64 & Sum + Max & 32 & 100 & BatchNorm    \\
AttentiveFP & $10^{-5}$ & 0.1 & ReLU &- & 3 & 64 & AttentiveFPReadout & 32 & 100 & -   \\
MPNN & $10^{-5}$ & 0.1 & ReLU & - & 6 & 64 & Sum + Max & 32 & 100 & -   \\
PAGTN & $10^{-5}$ & 0.1 & LeakyReLU &- & 5 & 64 & Weighted Sum + Max & 32 & 100 & -   \\
Graphormer & $10^{-5}$ & 0.1 & ReLU  & 8 & 1 & 64 & MaxPooling & 32 & 100  & LayerNorm  \\
\bottomrule
\end{tabular}
}
\end{table}



\section{Competing interests}
No competing interest is declared.

\section{Author contributions statement}

J.X. and T.F. conceived the experiment(s), J.X. conducted the experiment(s), and J.X., Y.Z., and T.F. analyzed the results. J.X., Y.Z., and T.F. wrote and reviewed the manuscript.

\section{Acknowledgments}
The authors thank the anonymous reviewers for their valuable suggestions. Y.Z. is partly supported by National Science Foundation (NSF) Award No. 2346158.

\bibliography{tmlr}
\bibliographystyle{tmlr}

\appendix
\section{Appendix}
You may include other additional sections here.

\end{document}

%% file: math_commands.tex
\usepackage{amsmath,amsfonts,bm}

\def\eqref#1{equation~\ref{#1}}

\def\1{\bm{1}}

\DeclareMathAlphabet{\mathsfit}{\encodingdefault}{\sfdefault}{m}{sl}
\SetMathAlphabet{\mathsfit}{bold}{\encodingdefault}{\sfdefault}{bx}{n}

